\documentclass[10pt,twocolumn,letterpaper]{article}

\usepackage[pagenumbers]{iccv}
\newcommand{\project}[1]{HouseTour}

\definecolor{iccvblue}{rgb}{0.21,0.49,0.74}
\usepackage[pagebackref,breaklinks,colorlinks,allcolors=iccvblue]{hyperref}
\usepackage[super]{nth}

\title{\project{}: A Virtual Real Estate A(I)gent}

\author{
Ata \c{C}elen$^{1,2}$
\quad
Marc Pollefeys$^{1,3}$
\quad
D\'aniel Bar\'ath$^{1,4}$
\quad
Iro Armeni$^{2}$\\
\textnormal{$^1$ETH Zürich \quad $^2$Stanford University \quad $^3$Microsoft Spatial AI Lab \quad $^4$HUN-REN SZTAKI}\\
\url{https://house-tour.github.io/}
}

\usepackage{multirow}
\usepackage[table]{xcolor}
\usepackage{makecell}
\usepackage{xr}
\usepackage[super]{nth}
\usepackage{tikz}
\usepackage{pgfplots} 
\pgfplotsset{compat=1.18}
\usepgfplotslibrary{groupplots}
\usepackage{url}
\usepackage[most]{tcolorbox}
\usepackage{float}

\newcommand{\slice}[5]{
  \pgfmathparse{0.5*#1+0.5*#2}
  \let\midangle\pgfmathresult

  \draw[thick,fill=#5!10] (0,0) -- (#1:1) arc (#1:#2:1) -- cycle;

  \node[label=\midangle:#4] at (\midangle:1) {};

  \pgfmathparse{min((#2-#1-10)/110*(-0.3),0)}
  \let\temp\pgfmathresult
  \pgfmathparse{max(\temp,-0.5) + 0.8}
  \let\innerpos\pgfmathresult
  \node at (\midangle:\innerpos) {#3};
}

\begin{document}

\twocolumn[{%
\renewcommand\twocolumn[1][]{#1}%
\maketitle
\vspace{-25pt}
\begin{center}
    \includegraphics[width=\textwidth]{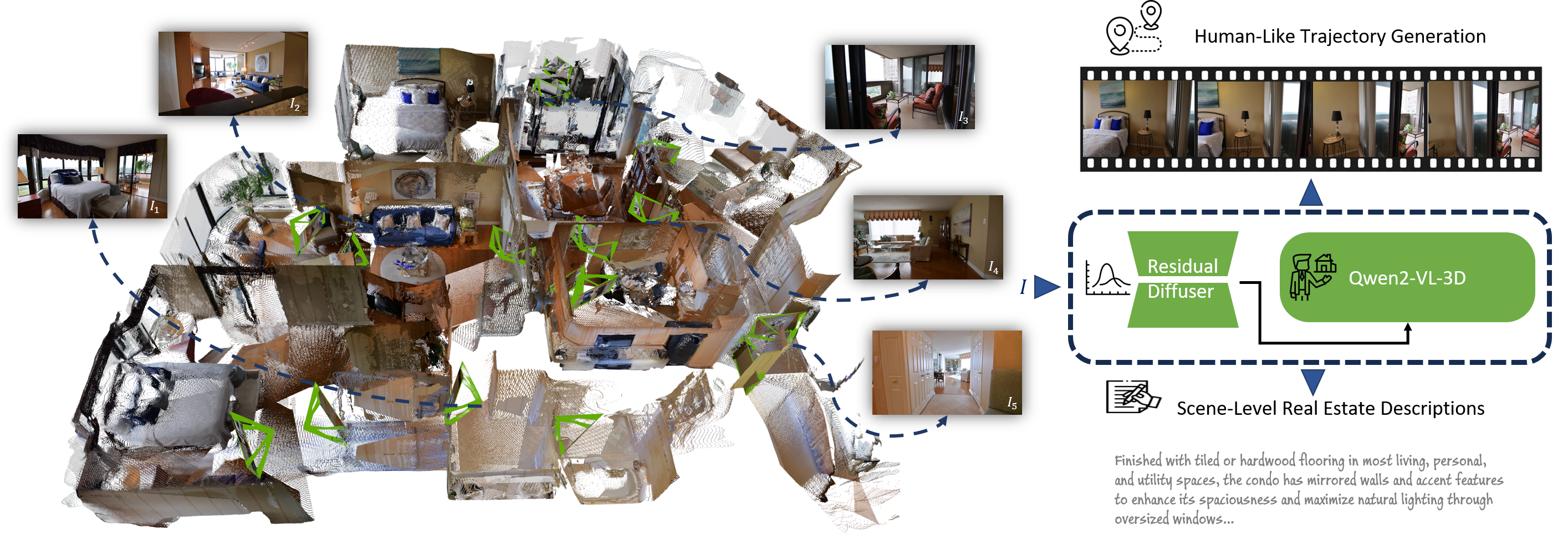}
    \captionof{figure}{\textbf{\project{}.} Given a set of images captured in an existing 3D space and their corresponding camera poses, our method tackles the task of 3D camera trajectory and textual summary generation. We focus on generating human-like trajectories and descriptions that can be used for real estate remote video tours of properties, thus highlighting spatial characteristics such as layout, functionality, architectural features, static building elements (e.g., appliances, windows, and doors), materials, and ambiance. To support the task, we present a novel dataset with real estate video tours, descriptions, and 3D reconstructions.}\label{fig:teaser}
\end{center}%
}]
\begin{abstract}
We introduce \project{}, a method for spatially-aware 3D camera trajectory and natural language summary generation from a collection of images depicting an existing 3D space. Unlike existing vision-language models (VLMs), which struggle with geometric reasoning, our approach generates smooth video trajectories via a diffusion process constrained by known camera poses and integrates this information into the VLM for 3D-grounded descriptions. We synthesize the final video using 3D Gaussian splatting to render novel views along the trajectory. To support this task, we present the \project{} dataset, which includes over 1,200 house-tour videos with camera poses, 3D reconstructions, and real estate descriptions. Experiments demonstrate that incorporating 3D camera trajectories into the text generation process improves performance over methods handling each task independently. We evaluate both individual and end-to-end performance, introducing a new joint metric. Our work enables automated, professional-quality video creation for real estate and touristic applications without requiring specialized expertise or equipment.
\end{abstract}

\vspace{-40 pt}
\section{Introduction}
\label{sec:intro}
Recent advances in vision-language models (VLMs) \cite{achiam2023gpt,wang2024qwen2,li2024llava} have enabled zero-shot generalization across a wide range of real-world image, video, and text applications, bridging the gap between curated research tasks and practical scenarios. However, generating videos grounded in an existing 3D space and describing spatial qualities in unstructured language---beyond merely listing contents---remains a challenge. 
This task requires geometric reasoning capabilities that current models lack. 

We present the novel task of spatially-aware 3D camera trajectory generation \textit{and} textual summarization from a collection of images depicting an existing 3D space. This task is closely related to creating house-tour videos, a popular format on YouTube, where over 624 million videos feature real estate agents and occupants showcasing their homes. This practice surged during the COVID-19 pandemic due to travel and interaction restrictions, providing renters and prospective buyers with remote access to properties. It remains a critical tool today in the U.S. real estate market, valued at 3.43 trillion dollars. However, providing such videos is labor-intensive, requiring expert real estate agents to visit properties with high-end videography equipment and manually craft detailed descriptions. Unlike scene captioning methods \cite{,chen2023end,yuan2022x,Chen2024SpatialVLM,Qi2025GPT4Scene}, these descriptions focus on spatial layout, functionality, architectural features, static building elements (\eg, appliances, windows, and doors), materials, and ambiance rather than simply enumerating furniture.

To address these challenges, we introduce \project{} (Fig.~\ref{fig:teaser}), a method to automatically generate house-tour videos from a set of captured images with known camera poses. Our approach enables users---without specialized expertise or equipment---to create professional-quality videos for real estate and touristic purposes. We extend an existing VLM \cite{wang2024qwen2} by fine-tuning it for 3D-grounded real estate descriptions. Our method generates smooth 3D camera trajectories using a diffusion process \cite{janner2022planning} constrained by the known camera poses and integrates this information directly into the VLM to ensure text alignment with the spatial path. To visualize the results, we synthesize the final video using 3D Gaussian splatting~\cite{Kerbl20233DGaussianSplats} to render novel views from the generated camera poses. We design the input to our method to accommodate practical use cases where typical end-users may struggle to capture smooth videos with smartphones. Additionally, using images instead of video enhances privacy by allowing selective content capture.

To support this task, we introduce the \project{} dataset, a curated collection of over 1,200 house tour videos featuring diverse properties ranging from apartments to multi-storey houses. Each video is accompanied by professionally captured smooth camera trajectories and real-estate-oriented textual descriptions, and generated 3D reconstructions. We compute ground-truth 3D camera pose information using off-the-shelf methods \cite{Schonberger2016SfM,Leroy2024MASt3R} and manually verify all dataset information to ensure accuracy, while removing visual and textual content that may infringe on privacy. Our dataset fills a gap in existing 3D visio-linguistic datasets \cite{achlioptas2020referit3d,chen2020scanrefer,jia2024sceneverse,zhang2023multi3drefer,hong20233d,ma2022sqa3d,anderson2018vision}, where camera trajectories are tailored to the 3D reconstruction task (close-up poses to object surfaces and jerky movements) and descriptions enumerate scene objects and their in-between relationships. 

Our contributions are threefold:\begin{itemize}
    \item We present a novel task: spatially-aware 3D camera trajectory and textual summary generation from a collection of  images, with the goal to resemble house tour videos. 
    \item We propose a new method, \project{}, that jointly models camera trajectory and language description generation, incorporating geometric constraints in the process.
    \item We release the \project{} dataset to support this task,  comprising house-tour videos with 3D reconstructions, and real-estate-style textual descriptions.
\end{itemize}

The dataset, code, and trained models publicly are available at \url{house-tour.github.io}.

\section{Related Work}
\label{sec:relwork}

\textbf{Long-Horizon Understanding and Captioning.} The Video-to-Text task traditionally emphasizes generating descriptive narratives from short video clips, focusing primarily on salient actions or prominent objects~\cite{Venugopalan2015Sequence,Zhou2018EndToEndVideoCaptioning}. Despite their effectiveness in general-purpose applications, these approaches typically lack detailed spatial reasoning and long-term contextual coherence, critical for describing spatial layouts and architectural details.  Recent advances ~\cite{Chen2024SpatialVLM,Qi2025GPT4Scene} explicitly model spatial relationships and semantic context to improve captioning. 
Nevertheless, these methods primarily operate on 3D scans or pre-recorded video sequences without dynamically integrating novel camera viewpoints or trajectories. 
This limits their utility for generating cohesive, spatially-aware narratives, \eg, for navigational videos, such as real estate tours.

Recent advances in 3D vision-language tasks recognize that real-world spatial context is crucial. Works like DenseCap~\cite{Johnson2016DenseCap} and Scan2Cap~\cite{Chen2021Scan2Cap} detect and describe objects within images or 3D scans, incorporating relational cues (\eg, “\textit{A couch next to a table}”). Furthermore, large-scale 3D datasets~\cite{Chang2017Matterport3D,Straub2019TheReplicaDataset,XiazGibsonEnv} have spurred research on embodied tasks (\eg, vision-and-language navigation~\cite{Anderson2018VLN}) where an agent must navigate and describe the environment. For instance, frameworks like EnvDrop~\cite{Tan2019EnvDrop} address navigation instructions grounded in real indoor scans, partially overlapping with our objective of context-aware commentary.

Lastly, only few Video-to-Text or Multi-Image-to-Text models can effectively handle very long sequences. Most models are designed for short-horizon videos that can be represented with a limited number of visual tokens \cite{zhang2023video, li2023videochat, wang2024tarsier, ataallah2024minigpt4}. TimeChat~\cite{ren2024timechat} notably extends the manageable video length by using a sliding window approach, but it lacks the foundational expertise required for architectural captioning of interior scenes. Only a handful of models meet both criteria—handling larger sequences of visual data while also incorporating domain knowledge for our task. Examples include Qwen2-VL~\cite{wang2024qwen2} (and its latest iteration, Qwen2.5-VL~\cite{bai2025qwen2}) as well as LLaVa-OneVision~\cite{li2024llava}.

\textbf{Trajectory Generation and Human-like Motion.}
Traditional camera trajectory estimation, fundamental to SLAM and SfM pipelines~\cite{MurArtal2015ORBSLAM,Engel2017DSO,Schonberger2016SfM}, focuses on accurately reconstructing camera poses from existing image sequences. Recent work, however, has explored generative approaches to trajectory planning, aiming to synthesize realistic camera movements that mimic human behavior or cinematic styles~\cite{Christie2008CameraControl,Koh2021Pathdreamer,Zhou2024LatentReframe}. These generative approaches typically utilize learned priors from human-recorded video datasets to produce plausible trajectories, but they often lack explicit geometric grounding, resulting in potential inaccuracies or physically infeasible paths. 

In robotics, existing literature often focuses on egocentric and allocentric motion as well as trajectory prediction~\cite{park2016egocentric, alahi2016social, li2020evolvegraph}, typically aiming to forecast short-term decisions given prior environment data. More recent methods such as SceneDiffuser~\cite{huang2023diffusion}, MotionDiffuser~\cite{jiang2023motiondiffuser}, and Decision Diffuser~\cite{ajay2022conditional} utilize diffusion-based models specifically designed for sequential decision-making tasks. In contrast, our approach, inspired by the Diffuser~\cite{janner2022planning}, leverages diffusion-based generative modeling explicitly conditioned on known 3D scene geometry. We formulate trajectory planning holistically rather than sequentially, enabling improved long-horizon decision-making. Such a strategy is particularly advantageous in our setting, where prior knowledge of the scene geometry is available, unlike many robotics scenarios that require simultaneous exploration and planning. While our technique draws on Diffuser’s inpainting-like paradigm for conditioning sparse observations, it mainly departs in two key ways. First, because our interaction spaces vary with different real estate layouts, we move away from learning an absolute trajectory in non-constant environments. Instead, we model human-like motion as a residual to spline interpolation, which proves more effective as a learning task. Second, we introduce a custom loss function tailored to our trajectory generation objectives, enhancing the quality of the resulting paths. 

\textbf{Datasets for Spatially-Aware Video Generation.}
Existing 3D datasets ~\cite{Chang2017Matterport3D,dai2017scannet,yadav2023habitat} predominantly feature environments captured using tripod systems or videos optimized for reconstruction purposes (\ie, staying close to object surfaces, lacking smooth movements, and failing to capture the entirety of rooms in single frames). In recent years, several datasets have extended these to connect 3D spaces with language \cite{achlioptas2020referit3d,chen2020scanrefer,jia2024sceneverse,zhang2023multi3drefer,hong20233d,ma2022sqa3d,anderson2018vision}, supporting tasks such as object referral, scene captioning, vision-language navigation, and reasoning. However, these datasets focus on describing furniture and object relationships while overlooking broader spatial aspects such as scene layout, architectural features, materials, and ambiance. They also lack real-world video trajectories designed to observe and highlight entire spaces and are not paired with professionally crafted textual narratives. Another related dataset, RealEstate10K \cite{zhou2018stereo}, consists of 7000 video snippets from online real estate YouTube videos, but these clips are significantly shorter than ours (1-10 seconds versus several minutes) and do not provide textual summaries of the scenes.
\section{\project{}}
\label{sec:methodology}

\begin{figure*}
    \includegraphics[width=\textwidth]{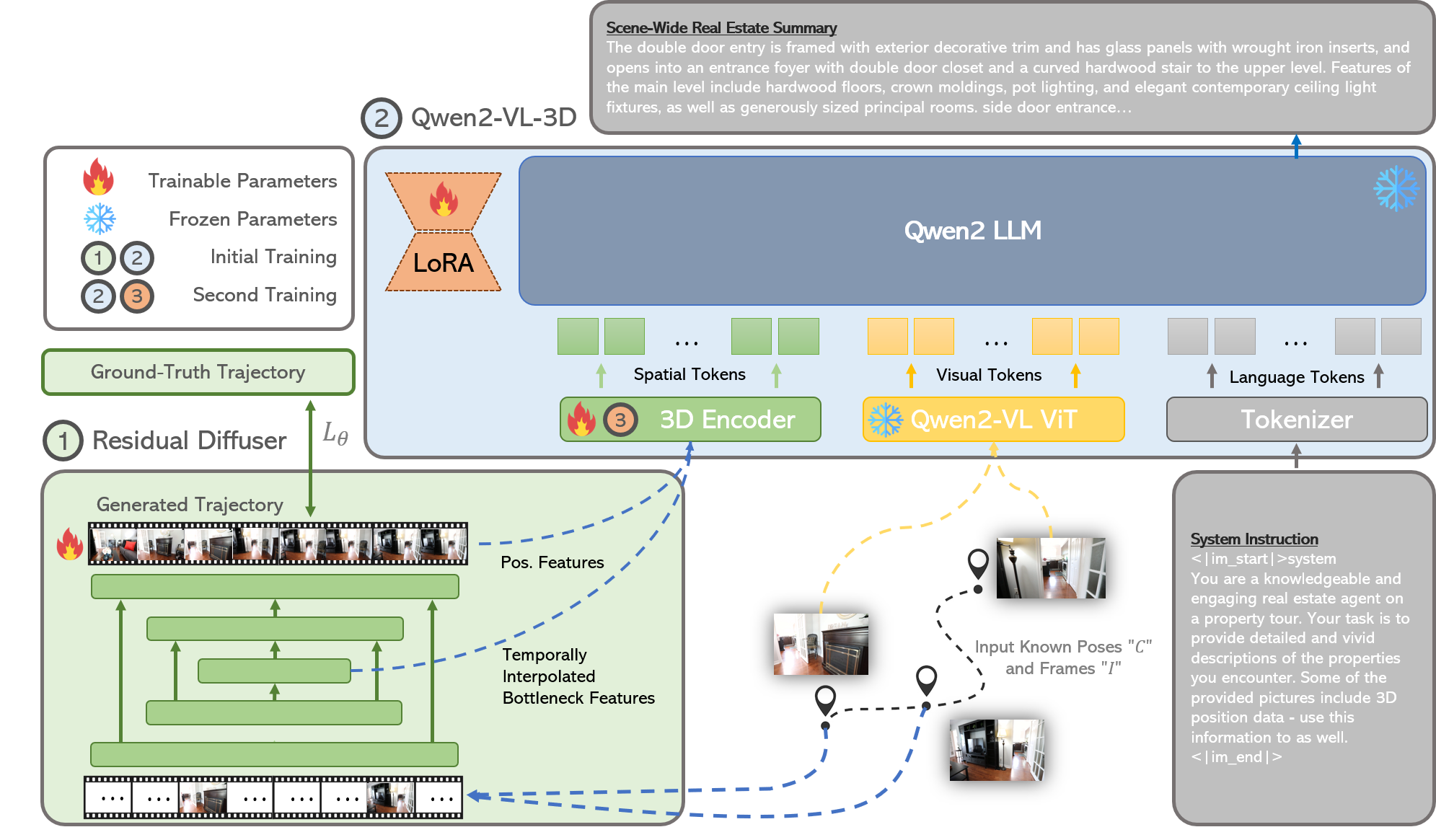}
    \captionof{figure}{\textbf{Pipeline Overview.} Given an prior tuple \((\mathcal{C}, I)\) of camera poses and images, our goal is to generate scene-level real estate summaries alongside a continuous camera trajectory that emulates human navigation. The sparse camera pose observations \(\mathcal{C}\) are refined and completed by the proposed \textit{Residual Diffuser} to obtain a smooth path. The resulting spatial features, along with the provided RGB frames \(I\), are then processed by the \textit{Qwen2-VL-3D} model to generate a coherent real estate summary.}\label{fig:architecture}
\end{figure*}

Given a tuple prior $(\mathcal{C}, \mathcal{I})$, where $\mathcal{C}=[c_1, c_2, ..., c_{N_c}]$ denotes a sequence of $N_c$ sparse, temporally ordered, and known camera poses and $\mathcal{I}$ the corresponding RGB frames for the camera poses, our objective is to generate a trajectory $\tau$ with $N > N_c$ frames and a scene-level summary $\boldsymbol{\Sigma}$ that emulate the motion and language narration of a professional real estate agent when touring a property. In short, the method takes posed images as input and returns (i) a continuous camera trajectory anchored to those observations and (ii) a descriptive summary of the scene.

The generated trajectory is represented by $\tau = [\boldsymbol{p}^1, \boldsymbol{p}^2, ... \boldsymbol{p}^N]$ and each camera pose $p_i = \{l_i, r_i\}$ along the trajectory is represented by a translation and a quaternion rotation vector, where $l_i = [x_i, y_i, z_i]$ and $r_i = a + b\mathbf{i} + c\mathbf{j} + d\mathbf{k}$. We assume prior spatial knowledge of the $\mathcal{C}$ camera poses. Specifically, each known camera pose is given by $c_i^{t_\tau} = \{l_i, r_i\}$ and $t_\tau$ represents the temporal order of the camera pose within the trajectory. Our generated summaries $\boldsymbol{\Sigma}$ are conditioned on both the sparse visual observations $\mathcal{I}$ and the spatial features $f_\theta(\mathcal{C})$ that are provided by our trajectory generation framework, \textit{Residual Diffuser}. In other words, the generation of $\boldsymbol{\Sigma}$ utilizes a tri-modal model--incorporating language, vision and 3D localization--referred to as \textit{Qwen2-VL-3D}. An overview of our method is in Figure \ref{fig:architecture}. For preliminaries see Supp.

\subsection{Diffusion-based Camera Trajectory Planning}
Our framework extends Diffuser \cite{janner2022planning} in terms of model architecture and inpainting-style conditioning approach. Analogous to image inpainting, where an incomplete image is denoised given a mask indicating known pixels, we denoise a trajectory conditioned on a set of sparsely known camera poses. Specifically, we treat $c_{1:N_c}$ as the ``mask" of known camera poses along the trajectory and denoise the rest of the trajectory around these poses. The fixed pose constraints are represented by a Dirac function~\cite{dirac1981principles} during the forward and reverse processes, allowing for deterministic sampling of these known camera poses at the predetermined timepoints $t_\tau$. Notably, our approach tackles trajectory planning purely within the ``observation" space, distinguishing it from Diffuser’s original formulation, which operates within a joint ``observation-action" space.

Additionally, Diffuser typically addresses a static interaction space, training the model to understand and plan within a fixed environment (such as navigating a constant maze with varying start and end points). In contrast, our scenario presents dynamically changing floor layouts, effectively creating a distinct maze to traverse in each iteration. This distinction makes the direct application of the Diffuser method effectively unsuitable for our case. The variability between scenes led us to change the formulation from learning absolute trajectory representations within each scene to learning residuals that mimic humane distinctions from spline interpolation. The spline solution is a robust initial approximation for moderately sparse observations.

The generated trajectory is calculated as: $\tilde{\boldsymbol{p}} = \mathcal{S} + \Delta\boldsymbol{p}$, where $\mathcal{S}$ is the interpolated spline for translations (SLERP for rotations~\cite{kremer2008quaternions}) between known camera poses and $\Delta\boldsymbol{p}$ denotes the predicted residuals. Under this formulation, for timesteps with known camera poses, we set the residual vector to zero. The reverse process for Residual Diffuser is:
\vspace{-5pt}
\begin{equation}
\label{eq:res_diff_reverse}
\begin{cases}
    \vec{0} = \delta(\boldsymbol{p}^i) & \text{ if } i \in t_\tau \\
    p_\theta(\Delta\boldsymbol{p}_{t-1}^i|\Delta\boldsymbol{p}_t^i, \mathcal{S}) = \mathcal{N}(\Delta\boldsymbol{p}_{t-1}^i; \mu_\theta, \Sigma_\theta)  & \text{ else} 
\end{cases}
\end{equation}

During the forward process, we again set the residuals of known camera poses to a zero vector, while the remaining residuals are diffused using the traditional approach outlined in Equation 2 in Supp. We train a U-Net architecture built with 1D convolutions (Figure \ref{fig:architecture} (1)), as in \cite{janner2022planning}, to predict the ground-truth residual from the diffused residual signal across uniformly sampled timesteps ranging from 1 to $T_{diff}$. Leveraging convolutions accommodates varying trajectory lengths during both training and inference. Furthermore, we significantly vary the sparseness of known camera poses during training for robust performance. The conditional diffusion process is as follows:
\vspace{-4pt}
\begin{equation}
\label{eq:res_diff_forward}
\begin{cases}
    \vec{0} = \delta(\boldsymbol{p}^i) & \text{ if } i \in t_\tau \\
    q(\Delta\boldsymbol{p}_{t}^i|\Delta\boldsymbol{p}_0^i, \mathcal{S}) = \mathcal{N}(\Delta\boldsymbol{p}_0^i; \sqrt{\alpha_t}x_0, (1-\bar{\alpha}_t)I)  & \text{ else} 
\end{cases}
\end{equation}

\noindent \textbf{Trajectory Loss.} In conventional denoising diffusion probabilistic model (DDPM) training, the loss is typically defined as the distance between the predicted and the ground-truth noise, a formulation that acts as a simplified variational bound~\cite{ho2020denoising}. However, for our application, directly minimizing the distance between the ground-truth and denoised camera poses may not be ideal for optimizing trajectories. This is partly because the ground-truth poses in our dataset were generated with 3D reconstruction and may contain inherent biases, such as denser sampling in low-texture areas and sparser sampling in high-texture regions. This could lead the model to learn undesirable patterns in order to minimize the objective. Moreover, a trajectory is a continuous function of camera poses that can be approximated by densely sampled points. We propose a loss function that adheres to these criteria and establishes a more effective distance measure between trajectories.

For a trajectory \(\tau\) of length \(N\), we define a spline segment (and SLERP for rotations) between each pair of consecutive camera poses, \(\boldsymbol{p}^{i}\) and \(\boldsymbol{p}^{i+1}\). We then efficiently evaluate this spline on \(n\) uniformly sampled points along each interval using Horner's Method \cite{horner1819xxi}, denoting the resulting set of points as:
\[
\mathcal{S}(\boldsymbol{p}^{i:i+1}) = \left(\boldsymbol{s}_1^{i:i+1}, \boldsymbol{s}_2^{i:i+1}, \dots, \boldsymbol{s}_n^{i:i+1}\right)
\]

Next, we compute the total Euclidean length of the trajectory \(\tau\) by summing the distances between successive camera poses. We define \(N_{eval} \gg N\) as the total number of evaluation points along the splines. These points are uniformly distributed along the trajectory based on Euclidean distance, with the precomputed spline values serving as indices for the evaluation points.

We compute the loss for translations using the \(L_2\) norm on uniformly sampled dense spline points. 
At the same time, for rotations, we employ a geodesic loss that provides a more suitable measure of distance on the \(\text{SO}(3)\) manifold. It is important to note that any residual vector added to a unit quaternion must be renormalized to ensure the quaternion remains of unit length. The resulting trajectory loss is formulated as follows:
\vspace{-4pt}
\begin{equation}
\label{eq:traj_loss}
    \mathcal{L}_\theta = \mathbb{E}_{t, \boldsymbol{\tau}, \boldsymbol{\epsilon}}\left[\| \boldsymbol{\epsilon_{pos}} - \boldsymbol{\epsilon}_\theta(pos_t, t)\|^2 + d_{geo}(\boldsymbol{\epsilon_{rot}}, \boldsymbol{\epsilon_{\theta}}(rot_t, t)\right]
\end{equation}

\subsection{Generating Real Estate Summaries}
Next, we generate real estate summaries that emulate professional house tours. These summaries emphasize the architectural features of the property rather than the objects visible in individual frames. Addressing this task requires a visual grasp of the vast architectural vocabulary; for example, differentiating among kitchen counter-top materials (quartz, ceramic, metal, wood, etc.) or recognizing various ceiling types (vaulted, cathedral, coffered, etc.). Moreover, this task demands the coherent processing of extensive, sparse multi-image data to accurately capture the complete layout of a property. Finally, we recognize that both visual and spatial information are essential to accurately locate each frame within the property and incorporate relevant spatial context into the summaries.

To this end, we integrate 3D spatial information as a third modality into a vision-language model, namely \textit{Qwen2-VL-3D}, leveraging the spatial features provided by the Residual Diffuser to enhance summary quality (Figure \ref{fig:architecture} (2)). Our training builds on the Qwen2-VL~\cite{wang2024qwen2} model, which serves as a robust foundation. Unlike many vision-language models that struggle to process large batches of images, Qwen2-VL provides a rich visual knowledge base that effectively connects with language and can attend to details across a large amount of visual tokens. 

In the first step of training, we employ the parameter-efficient LoRA~\cite{hu2022lora} fine-tuning method to train the Qwen2-VL model on the task of generating real-estate summaries. For this, we uniformly sample \(N_{\text{frames}}\) frames from each house tour video to serve as visual cues during multi-image training. We set \(N_{\text{frames}}\) to 96, which is chosen based on the memory requirements of training while ensuring sufficient scene coverage. This step ensures that the fine-tuned version effectively captures the language style of house tour summaries and incorporates the appropriate architectural terminology in its generated output.

In the second stage, we integrate spatial understanding into the summary generation process. First, we add the special tokens \(\texttt{<|traj\_start|>}\), \(\texttt{<|traj\_pad|>}\), and \(\texttt{<|traj\_end|>}\) to the VLM's vocabulary. The start and end tokens define the boundaries where spatial features are inserted into the user prompt, and the pad token is replaced by the corresponding spatial token. Each spatial token is provided alongside the visual tokens from the corresponding scene location, though the visual tokens are not required to be paired with spatial tokens in return. This setup offers the flexibility to include frames with and without spatial features during training and inference.

Next, we build the adapter that transforms the raw features and absolute positional information coming from the Residual Diffuser into token representations compatible with the language processing components of the Qwen2-VL model. For each frame fed into the VLM, we denoise its corresponding pose $\boldsymbol{p}^i_0$. We also get the temporally downsampled features from the bottleneck layer of the Residual Diffuser and upsample it with interpolation. We concatenate $\boldsymbol{p}^i_0$ with $\boldsymbol{f}^i_0$, where $\boldsymbol{f}^i_0$ corresponds to the bottleneck feature from the last step of trajectory denoising. Utilizing the bottleneck layer features along with the denoised camera pose information ensures a high-level global representation of the trajectory. Lastly, the concatenated spatial features are passed through a linear layer, which maps them into the embedding space of Qwen2-VL's language component. We use a single token to encode each frame’s spatial information. Implementation details are in Supp.

\subsection{House Tour Videos via 3D Gaussian Splatting}
To assess the visual quality of the generated trajectories, we train a Gaussian Splat \cite{Kerbl20233DGaussianSplats} using all the ground-truth poses along with the reconstructed point clouds. We then render the camera poses, denoised by the Residual Diffuser, using only the sparse views. These videos are solely for visualizing the trajectories; during inference, the end user will have access only to the sparse views. While recent studies~\cite{xu2024depthsplat, chen2025mvsplat360, fan2024instantsplat} address synthesizing scenes from sparse views, this problem falls outside the scope of our work.

\section{\project{} Dataset}
\label{sec:dataset}
We introduce the \project{} dataset, which features scene-scale human trajectories, dense point clouds, and real estate descriptions, all derived from in-the-wild RGB real estate tour videos. The data is procured from professional real-estate agencies. Our dataset comprises 1639 videos showcasing properties ranging from condos to multi-storey apartments. Of these, 1298 videos are transcribed---half with timestamped descriptions---capturing the detailed professional language used by real estate agents use for both interior and exterior spaces. Additionally, we provide 3D reconstructions for 878 scenes, while the remaining scenes experienced partial or complete reconstruction failures. Further details on the reconstruction pipeline, the dataset creation process, and statistics are in Supp.
\section{Experiments}
\label{sec:experiments}
We evaluate \project{} (\emph{Residual Diffuser + Qwen2-VL-3D}) on human‐like trajectory generation and multi‐image scene summarization. In Section \ref{sec:end2end}, we measure end-to-end performance with a novel joint metric and compare to the best performing baselines per task. We also provide per task analysis in Sections \ref{sec:traj_results}-\ref{sec:text_results}. All experiments are evaluated on the test set of our \project{} Dataset.

\subsection{Evaluation Metrics}
\noindent \textbf{Trajectory Generation.}
We report the recall‐based metrics, \emph{R@50cm}, \emph{R@75cm}, and \emph{R@1m}, which indicate the percentage of predictions with a translation error less than 50cm, 7cm, or 1m respectively. We then evaluate the trajectories using several distance‐ and shape‐based metrics: \emph{Euclidean Distance}, \emph{Dynamic Time Warping} (DTW), \emph{Hausdorff Distance}, \emph{Fréchet Distance}, and the \emph{L\(_2\) Chamfer Distance}. We examine rotational quality via \emph{Quaternion Distance} and \textit{Geodesic Distance}; and use peak-signal-to-noise (\emph{PSNR}) and structural similarity (\emph{SSIM}) to evaluate the rendering performance on the 3D Gaussian splatting output.

\vspace{2pt}
\noindent \textbf{Scene Summarization.}
We organize our metric selection around two primary goals: stylistic and factual alignment. For stylistic alignment, we use BLEU (B)~\cite{papineni2002bleu}, ROUGE‐L (ROU‐L)~\cite{lin2004rouge}, METEOR (MTR)~\cite{banerjee2005meteor}, and CIDEr (CDr)~\cite{vedantam2015cider}. These $n$-gram based metrics are useful for assessing style, vocabulary, and syntax but cannot adequately capture aspects of factual correctness, coherence, or hallucinations---considerations that are critical for producing coherent summaries. To address these limitations, we adopt a commonly used preference‐based evaluation approach: Bradley–Terry scores (BT). By comparing pairs of generated texts, this method captures their overall quality, including factuality and overall coherence. Building on recent approaches that use Bradley-Terry models to rank performance and employ LLMs as judges based on their own preferences~\cite{chiang2024chatbotarenaopenplatform, gao2023human}, we leverage GPT-4o~\cite{hurst2024gpt} to compare the summaries produced by each method. In each iteration, the LLM is presented with the ground-truth summary alongside two generated summaries from different methods and is tasked with selecting the one that most closely matches the ground truth. To ensure fairness, the order that the summaries are presented to the LLM is randomized. For more details see the supplementary material.

\vspace{2pt}
\noindent \textbf{End-To-End Performance.}
To measure end-to-end performance, we develop a new metric, the \emph{Spatio-Linguistic} 
score (\textbf{SLS}), that measures the joint performance of methods on the tasks of 3D camera trajectory and textual summary generation. More specifically, its role is to evaluate spatial geometry (translation and rotation) with respect to the ground truth trajectory and linguistic overlap with professionally annotated real estate summaries. It is computed as the harmonic mean of the Translation Recall at 75cm (R@75cm), the Rotation Score (Rot. Score), and the Bradley--Terry score (BT) , and ranges from 0 to 100. Here, the Rotation Score is defined as \(1 - \frac{\text{geo. dist.}}{\pi}\). The first two metrics evaluate trajectory and the latter summary generation. Together, they offer a comprehensive view of performance on the joint task. \textit{Why R@75cm?} As shown in Table \ref{tab:traj}, our approach outperforms the baselines at larger recall thresholds but loses its comparative edge for tighter thresholds. We attribute this pattern to the increase in uncertainty as the distance to the closest known pose increases. Interpolation-based methods have high representation power where the evaluation points are close to the known poses due to the continuity and smoothness of trajectory curves. As the distance to the closest known pose grows, the uncertainty around trajectory prediction increases, causing interpolation‐based methods to suffer greater performance degradation compared to our approach. Therefore, we select a \textit{75cm} threshold as a practical balance between tightness and representational power.

\subsection{End-to-End Performance}
\label{sec:end2end}
In Table \ref{tab:joint_table}, we compare the end-to-end performance of our method with a composed baseline. Due to the absence of a method that can solve this joint task, we devise one by using the best performing methods per task: Catmull-Rom Spline~\cite{twigg2003catmull} for camera trajectory generation and Qwen2-VL-7B (SFT) for textual summary generation---a finetuned version of Qwen2-VL \cite{wang2024qwen2} for the task of generating \textit{real-estate} summaries. \project{} outperforms the baseline on all metrics, including the joint one. 

\begin{table}[htbp!]
\centering
\resizebox{\columnwidth}{!}{
\begin{tabular}{ccccc}
\toprule
\textbf{Methods} & \textbf{R}@75cm~$\uparrow$ & \textbf{Rot. Score}~$\uparrow$ & \textbf{BT}~$\uparrow$ & \textbf{SLS}~$\uparrow$ \\
\midrule
Baseline & 57.1 & 96.8 & 71.4 & 71.7 \\
\project{} & \cellcolor{green}\textbf{60.2} & \cellcolor{green}\textbf{97.1} & \cellcolor{green}\textbf{79.5} & \cellcolor{green}\textbf{76.0} \\
\bottomrule
\end{tabular}%
}
\caption{\textbf{End-to-End Performance on 3D camera trajectory and textual summary generation.} The baseline method consists of Catmull-Rom Spline~\cite{catmull1974class} and Qwen2-VL-7B (SFT), and \project{} (ours) of Residual Diffuser and Qwen2-VL-3D.
}
\label{tab:joint_table}
\end{table}

\subsection{Trajectory Generation}
\label{sec:traj_results}
\begin{table*}[htbp!]
\resizebox{\textwidth}{!}{%
\begin{tabular}{c cc c cccccccc}

\toprule
\multirow{2}*{\textbf{Methods}} &
\multicolumn{7}{c}{\makecell{\textbf{Translation}}} & 
\multicolumn{2}{c}{\makecell{\textbf{Rotation}}} &
\multicolumn{2}{c}{\makecell{\textbf{Rendering}}} \\

& \textit{R@50cm}~$\uparrow$ & \textit{R@1m}~$\uparrow$ & \textit{Euclidean Dist.}~$\downarrow$ & \textit{DTW}~$\downarrow$ & \textit{Hausdorff Dist.}~$\downarrow$ & \textit{Frechet Dist.}~$\downarrow$ & \textit{Chamfer Dist.}~$\downarrow$ & \textit{Quaternion Dist.}~$\downarrow$ & \textit{Geodesic Dist.}~$\downarrow$ & \textit{PSNR}~$\uparrow$ & 
\textit{SSIM}~$\uparrow$ \\ 
\midrule
Lin. Interp. & 41.2\% & 59.8\% & 145.8 & 192.1 & 118.7 & 126.7 & 109.5  & 0.0432 & 0.20 & 14.20 & \cellcolor{green}\textbf{0.557} \\
Catmull-Rom & 45.9\% & 64.7\% & 106.2 & 146.3 & 89.3 & 95.8 & 83.4 & 0.0079 & 0.10 & 14.22 & \cellcolor{green}\textbf{0.557} \\
Residual Diffuser (Ours) & \cellcolor{green}\textbf{46.2\%} & \cellcolor{green}\textbf{69.4\%} & \cellcolor{green}\textbf{73.9} & \cellcolor{green}\textbf{128.8} & \cellcolor{green}\textbf{76.3} & \cellcolor{green}\textbf{81.2} & \cellcolor{green}\textbf{75.5} & \cellcolor{green}\textbf{0.0073} & \cellcolor{green}\textbf{0.09} & \cellcolor{green}\textbf{14.24} & 0.556 \\

\bottomrule
\end{tabular}%
}
\caption{\textbf{Spatial Trajectory Generation Performance.} We compare our generated trajectories to interpolation‐based baselines that do not account for “human‐like” motion. \textit{Translation} performance is reported in \textbf{cm}, \textit{Quaternion Distance} is measured as the Euclidean distance between unit quaternions, and \textit{Geodesic Distance} is reported in \textbf{radians}.}
\vspace{-10pt}
\label{tab:traj}
\end{table*}
Table \ref{tab:traj} presents a comparison of our \textit{Residual Diffuser} against two interpolation-based baselines, \textit{Linear Interpolation} and \textit{Catmull-Rom Splines}, for trajectory generation. For rotational data, we adjust the Linear Interpolation baseline to linearly interpolate quaternions, while the Catmull–Rom one uses spherical linear interpolation (SLERP). We construct our spline-based baseline using the Catmull-Rom spline because, unlike other spline variations such as the B-Spline\cite{de1972calculating}, it passes through all the control points.

In order to evaluate methods on cases with varying number of frames that range from sparser to denser coverage of the property, we vary the frequency of the known camera poses between every $5^{th}$ and $15^{th}$ frame of the ground-truth video. Our method outperforms baselines across all translation and rotation metrics. In particular, the highest recall for \textit{R@1m} indicates that our method incurs fewer ``large–errors”—errors substantial enough to be considered major misalignments—than the interpolation methods. In addition, significantly reduced distance metrics in translation, along with lower quaternion and geodesic distances in rotation ,imply more accurate pose generation. For rendering-based metrics, PSNR and SSIM remain comparable to baseline values; these metrics are less sensitive to slight pose differences—especially when the rendered image content (\eg, lighting, texture, and overall scene structure) is preserved and, thus, are less informative on the quality of the final video.

\begin{figure}[htp]
        \includegraphics[width=\columnwidth]{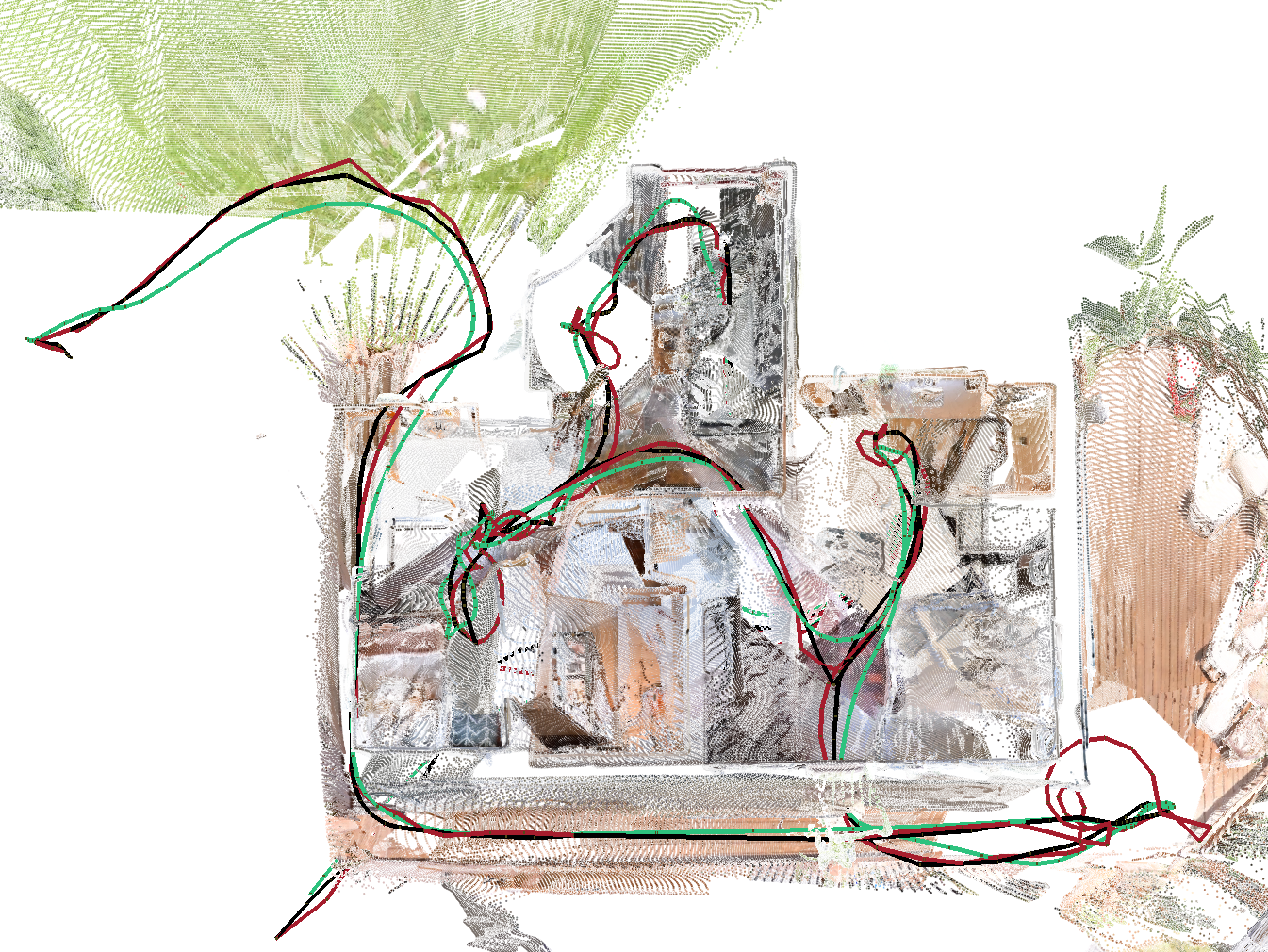}%
    \caption{\textbf{Trajectory Visualization Within the 3D Reconstructions (top view).} Our method, Residual Diffuser, achieves a more human-like and smooth trajectory than the baseline Catmull-Rom Spline. \textbf{Black:} Ground-Truth, \textbf{Green:} \textit{Residual Diffuser} and \textbf{Red:} Catmull-Rom Spline.}
    \label{fig:trajquals}
\end{figure}

These results confirm that incorporating data-driven diffusion into trajectory prediction better captures human-like motion tendencies and yields more robust and precise results than purely geometric interpolation. Figure \ref{fig:trajquals} further supports this, showcasing more human-like and smooth camera trajectories. For more results see supp.

\subsection{Scene Summarization}
\label{sec:text_results}
In Table \ref{tab:summary}, we compare our \emph{Qwen2-VL-3D} method with zero‐shot and supervised fine‐tuned variants of foundation VLMs. Since many state‐of‐the‐art VLMs struggle when faced with a large set of input images, thus limiting their scene‐level understanding, we use \emph{LLaVa-OneVision-7b} \cite{li2024llava} and \emph{Qwen2-VL-7b} \cite{wang2024qwen2} as zero-shot baselines, which have a demonstrated robustness to larger multi‐image inputs. We also finetune Qwen2-VL-7b on our dataset without any interactions with the trajectory generation to learn the real-estate language style (Qwen2-VL-7b (SFT)).

\begin{table}[htbp!]
\centering
\resizebox{\columnwidth}{!}{
\begin{tabular}{c cc c ccccccccc} 
\toprule
\multirow{2}{*}{Methods} &
\multicolumn{8}{c}{\makecell{\textbf{Multi-image Scene Summarization}}} \\

& B1~$\uparrow$ & B2~$\uparrow$ & B3~$\uparrow$ & B4~$\uparrow$ & ROU-L~$\uparrow$ & MTR~$\uparrow$ & CDr~$\uparrow$ & BT~$\uparrow$\\ 
\midrule
LLaVa-OneVision-7b & 0.259 & 0.128 & 0.055 & 0.024 & 0.189 & 0.109 & 0.001 & 0.04 \\ 
Qwen2-VL-7B & 0.272 & 0.137 & 0.064 & 0.029 & 0.195 & 0.13 & 0.005 & 0.04\\ 
Qwen2-VL-7B (SFT) & 0.363 & 0.220 & 0.126 & 0.070 & 0.231 & 0.175 & \cellcolor{green}\textbf{0.026} & 0.71 \\ 
\midrule

Qwen2-VL-3D (Ours) & \cellcolor{green}\textbf{0.433} & \cellcolor{green}\textbf{0.264} & \cellcolor{green}\textbf{0.154} & \cellcolor{green}\textbf{0.090} & \cellcolor{green}\textbf{0.24} & \cellcolor{green}\textbf{0.193} & 0.021 & \cellcolor{green}\textbf{0.79} \\ 

\bottomrule
\end{tabular}%
}
\caption{\textbf{Linguistic Summary Evaluation.} We compare \emph{Qwen2-VL-3D} against other foundation VLM models. \emph{(SFT)} denotes that the model has been fine-tuned on our dataset without any input from the trajectory generation.}
\vspace{-10pt}
\label{tab:summary}
\end{table}
\begin{figure*}
    \includegraphics[width=\textwidth]{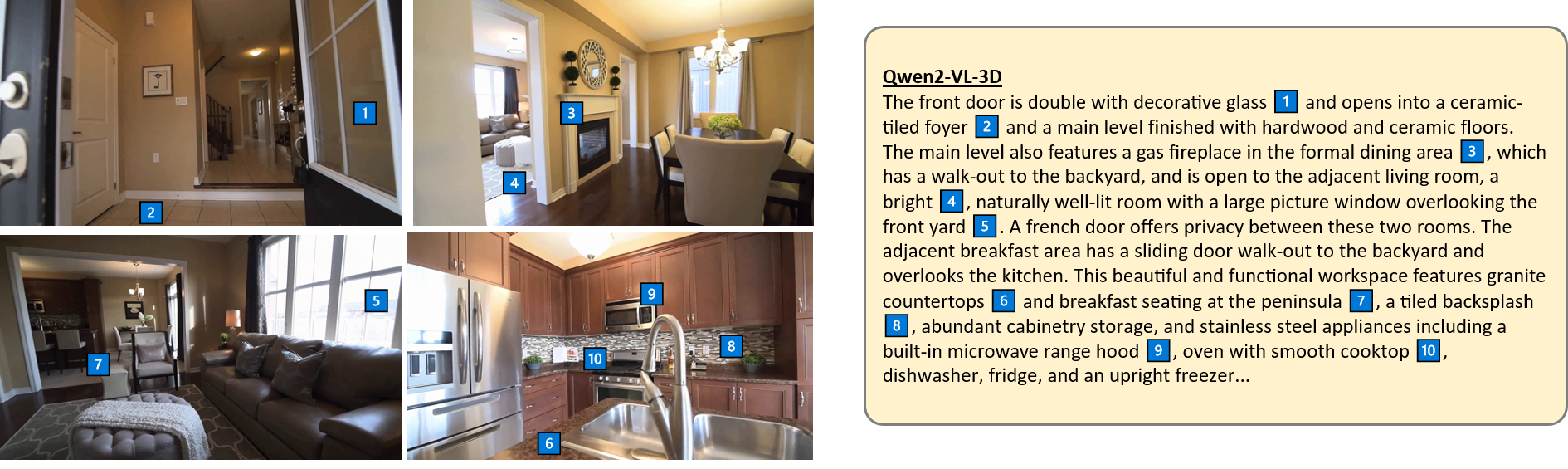}
    \captionof{figure}{\textbf{Qualitative Results for Scene-Level Summary Generation.} Our method creates accurate real-estate descriptions that capture the architectural style and elements of the space. We showcase an example of our summary generation, including sampled images of the space and a mapping between generated text and spatial elements to facilitate the reader.} \label{fig:summary_quals}
\end{figure*}

By incorporating 3D‐aware information, the resulting multi‐image scene summaries are noticeably more coherent and descriptive, as evidenced by improvements across most \(n\)-gram metrics and preference‐based evaluations. Fine‐tuning also has a pronounced effect on preference scores, as the fine‐tuned model decisively outperforms its zero‐shot counterpart. While a performance gain is unsurprising, the substantial gap underscores the limitations of off‐the‐shelf VLMs for generating more applied language styles.

Figure \ref{fig:summary_quals} shows an example of the summaries generated by our method. As shown, \project{} can accurately identify the layout, materials, architectural elements, and ambiance. For more results, including a user study, see supp.

\begin{table*}[htbp!]
\resizebox{\textwidth}{!}{
\begin{tabular}{c cc c cccccccc}

\toprule
\textbf{Pose Frequency} &
\multirow{2}*{\textbf{Methods}} &
\multicolumn{5}{c}{\makecell{\textbf{Translation (cm)}}} & 
\multicolumn{2}{c}{\makecell{\textbf{Rotation}}} \\

\textit{(every N frame)} &
& \textit{Euclidean Dist.}~$\downarrow$ & \textit{DTW}~$\downarrow$ & \textit{Hausdorff Dist.}~$\downarrow$ & \textit{Frechet Dist.}~$\downarrow$ & \textit{Chamfer Dist.}~$\downarrow$ & \textit{Quaternion Dist.}~$\downarrow$ & \textit{Geodesic Err.}~$\downarrow$ \\ 
\midrule
\multirow{3}*{\nth{5}}
& Lin. Interp. & 87.5 & 151.6 & 95.4 & 99.9 & 89.6 & 0.0368 & 0.19 \\
& Catmull-Rom & 56.0 & 103.5 & 64.2 & 67.4 & 61.5 & 0.0053 & \cellcolor{green}\textbf{0.07} \\
& Residual Diffuser (Ours) & \cellcolor{green}\textbf{41.5} & \cellcolor{green}\textbf{96.3} & \cellcolor{green}\textbf{57.8} & \cellcolor{green}\textbf{60.3} & \cellcolor{green}\textbf{58.4} & \cellcolor{green}\textbf{0.0052} & \cellcolor{green}\textbf{0.07} \\
\midrule
\multirow{3}*{\nth{10}}
& Lin. Interp. & 268.3 & 278.1 & 166.3 & 182.2 & 149.5 & 0.0714 & 0.33 \\
& Catmull-Rom & 219.0 & 238.4 & 140.6 & 154.9 & 128.6 & 0.0133 & 0.14 \\
& Residual Diffuser (Ours) & \cellcolor{green}\textbf{151.1} & \cellcolor{green}\textbf{197.3} & \cellcolor{green}\textbf{114.8} & \cellcolor{green}\textbf{125.2} & \cellcolor{green}\textbf{110.7} & \cellcolor{green}\textbf{0.0131} & \cellcolor{green}\textbf{0.13} \\
\midrule
\multirow{3}*{\nth{15}}\
& Lin. Interp. & 559.9 & 414.9 & 235.9 & 272.9 & 202.4 & 0.1059 & 0.46 \\
& Catmull-Rom & 510.9 & 388.5 & 217.3 & 254.7 & 189.6 & 0.0477 & 0.28 \\
& Residual Diffuser (Ours) & \cellcolor{green}\textbf{371.2} & \cellcolor{green}\textbf{321.1} & \cellcolor{green}\textbf{177.1} & \cellcolor{green}\textbf{206.6} & \cellcolor{green}\textbf{163.9} & \cellcolor{green}\textbf{0.0471} & \cellcolor{green}\textbf{0.27} \\

\bottomrule
\end{tabular}
}
\caption{\textbf{Ablation Study} on the impact on trajectory generation performance when varying the frequency of the known camera poses.}
\vspace{-10pt}
\label{tab:ablation_traj_freq}
\end{table*}

\subsection{Ablation Studies}

Table \ref{tab:ablation_traj_freq} examines how the frequency of observations impacts trajectory generation. In this experiment, observations are uniformly sampled every 5th, 10th, and 15th frame, and the corresponding performance metrics are reported. As anticipated, sparser sampling provides less information, which results in higher errors for all methods. Our findings indicate that our method performs reliably in both dense and sparse scenarios, establishing it as an overall better choice. Notably, the table reveals that at a moderate sampling frequency (every 10th frame) our method works the best. The Euclidean distance error is reduced by 32\% compared to the closest baseline with moderate sampling frequency, while reductions of approximately 28\% are observed when sampling every 5th and 15th frames.

\begin{table}[htbp!]
\centering
\begin{tabular}{ccc} 
\toprule
\textbf{Methods} & \textbf{w/o 3D Pos.} & \textbf{w/ 3D Pos.} \\
\midrule
Qwen2-VL-7B (SFT) & 44\% & 32.5\% \\
Qwen2-VL-3D (Ours) & \cellcolor{green}\textbf{56\%} & \cellcolor{green}\textbf{67.5\%} \\

\bottomrule
\end{tabular}

\caption{\textbf{Ablation Study} on the impact of 3D positional information on the summarization performance, using Bradley--Terry probabilities. The values denote the \textbf{Win Percentages}.}
\label{tab:ablation_3d}
\vspace{-10pt}
\end{table}
In Table \ref{tab:ablation_3d}, we investigate the effectiveness of the presence of 3D information for the summary generation task. For this purpose, we partition our test set into scenes that include 3D information and those that do not. The results indicate that when 3D information is available, \textit{Qwen2-VL-3D} is significantly favored over the finetuned Qwen2-VL-7B model, highlighting the value of spatial priors in summary generation. Additionally, the performance boost that our method achieves in the absence of 3D data suggests that the model may, to some extent, be learning to ``localize" the images within the scene during training. As a reminder, during training, Qwen2-VL-3D receives as input data with and without 3D positioning.

Results on out-of-distribution scenes are in the supp.

\section{Conclusion}

We introduced \project{}, a method for spatially-aware 3D camera trajectory and textual summary generation from a collection of images. Our approach addresses the limitations of existing VLMs by incorporating geometric reasoning through a diffusion process, enabling the creation of realistic house-tour videos without specialized equipment or expertise. We also presented the \project{} dataset, which uniquely combines real-world house-tour videos, accurate 3D reconstructions, and professionally crafted textual descriptions, facilitating comprehensive evaluation of spatially-aware video methods. Future works can explore incorporating information from VLMs to jointly guide the trajectory diffusion process and the development of Gaussian splatting methods that can fill the gaps between images without generating non-existing content.

{
    \small
    \bibliographystyle{ieeenat_fullname}
    \bibliography{main}
}

\newpage

\begin{center}
\bfseries\Large
\project{}: A Virtual Real Estate A(I)gent:
Appendix
\end{center}

\begin{abstract}
In the supplementary material, we provide:
\begin{enumerate}
\item Details on the \project{} dataset (Sec. \ref{sec:supp_dataset})
\item More qualitative results (Sec. \ref{sec:supp_quals})
\item User Evaluation of Text Generation (Sec. \ref{sec:user_eval}) 
\item Implementation details (Sec. \ref{sec:supp_implement})
\item Out-of-Distribution Scenes (Sec. \ref{sec:ood_scenes})
\item Details on the Metric Scale Evaluation (Sec. \ref{sec:metric_eval})
\item Details on the Bradley-Terry evaluation (Sec. \ref{sec:supp_BT})
\item Recall Plots for the trajectory generation (Sec. \ref{sec:supp_recall})
\item Preliminary information on diffusion (Sec. \ref{sec:supp_prelim})

\end{enumerate}
\end{abstract}

\section{\project{} Dataset}
\label{sec:supp_dataset}

\subsection{Creation Details}
At the start of our reconstruction pipeline, we select a subset of video frames for use in the process. Since our videos range from a few minutes to 15 minutes in length, using all frames would be computationally impractical. Our objective is to choose a minimal yet effective set of keyframes that maintain significant overlap to ensure accurate reconstruction. To accomplish this, we use an algorithmic approach that evaluates factors such as optical flow and keypoint matches between consecutive views.

We then trim the beginning and end of the selected keyframe sequence to remove frames that induce spatial jumps—such as exterior drone shots, which could largely affect the 3D reconstruction. To accomplish this, we use an off-the-shelf vision-language model, BLIP2 \cite{li2023blip}, to classify the keyframes at the sequence boundaries as exterior shots. 

For the 3D scene reconstruction, we employ the COLMAP~\cite{Schonberger2016SfM} structure-from-motion approach. We generate image pairs from the keyframes leveraging their inherent sequential order and augment these pairs with additional ones identified through traditional image retrieval techniques~\cite{schonberger2017vote} to simulate loop closure during the reconstruction process. We perform dense 2D-to-2D matching between paired frames using Mast3r~\cite{Leroy2024MASt3R} and subsequently map them with COLMAP after geometrically verifying the pixel correspondences.

Lastly, if a video contains speech, we use Whisper~\cite{radford2023robust} to extract the transcriptions along with their timestamps. If there is no speech, we obtain video descriptions as they stylistically align with the transcribed scene descriptions. To protect privacy, we employ GPT-4o to automatically filter out sensitive details such as addresses, personal names, and phone numbers from the video transcriptions. Additionally, we manually edit sections that mention neighborhood information or amenities, since such details are not visually represented in the videos.

\subsection{Dataset Statistics}
\paragraph{3D Reconstruction and Scene-Level Descriptions.}
The 3D reconstruction process for a single scene can take up to 40 hours, depending on the number of keyframes extracted from the videos. Reconstructing over 1,600 scenes may require 3 to 4 months of CPU time. Additionally, the keyframe extraction and matching processes are GPU-intensive. To manage this, we employ a high-performance computing cluster for dataset acquisition. Collecting the dataset typically takes about 7 days when using a job array with 20 parallel jobs. Each job requires 48GB of CPU memory and a 32GB NVIDIA Tesla V100 GPU. To handle extremely long runtimes, we limit each job to a maximum duration of 40 hours.\newline

\begin{table}[htbp!]
\hspace{1cm}
\begin{tikzpicture}[scale=2]

\newcounter{a}
\newcounter{b}
\foreach \p/\t/\c in {60/SUCCESS/green, 9/FAILED/red, 9/MISALIGNED/yellow, 15/BAD SCALING/orange, 7/INCOMPLETE/black}
  {
    \setcounter{a}{\value{b}}
    \addtocounter{b}{\p}
    \slice{\thea/100*360}
          {\theb/100*360}
          {\p\%}{\t}{\c}
  }

\end{tikzpicture}
\caption{\textbf{Breakdown of the reconstruction outcomes.}}
\label{tab:recon_breakdown}
\end{table}

Table \ref{tab:recon_breakdown} provides a detailed analysis of our pipeline's reconstruction performance. Scenes marked as \textit{SUCCESS} have been reconstructed successfully. \textit{INCOMPLETE} refers to reconstructions that failed to register the entire scene due to tracking loss, often caused by textureless areas or the lack of complete covisibility graph data. \textit{MISALIGNED} scenes have errors in reconstruction leading to incorrect rotations of some scene portions. Scenes labeled with \textit{BAD SCALING} have errors resulting in discrepancies in scale across parts of the output model. We classify scenes as \textit{FAILED} if they exhibit multiple of the aforementioned issues or have significant errors in the final model.

Each reconstructed scene includes a dense point cloud with more than one million vertices, 2D-to-3D correspondences, and outputs from COLMAP \cite{Schonberger2016SfM}, including images, camera data, and 3D points in binary files. Additionally, the selected keyframes and their timestamps are listed. For scenes with descriptions, we provide either a CSV file containing text and timestamp information or a plain text file with the description.

\begin{table}[htbp!]
\hspace{1cm}
\begin{tikzpicture}[scale=2]

\foreach \p/\t/\c in {63/Transcripts/green, 37/Descriptions/red}
  {
    \setcounter{a}{\value{b}}
    \addtocounter{b}{\p}
    \slice{\thea/100*360}
          {\theb/100*360}
          {\p\%}{\t}{\c}
  }

\end{tikzpicture}
\caption{\textbf{Distribution of summary types in the \project{} dataset.} Transcripts include timestamped information.}
\label{tab:descriptions}
\end{table}
Furthermore, we retrieved real estate descriptions for 1,298 out of 1,639 scenes. Of these, 813 descriptions were transcribed using the Whisper model, while the remaining ones were sourced from descriptions. We include all scenes in the dataset, even if they do not have 3D reconstructions or summaries. We utilize the scenes without 3D reconstruction as additional data to the VLM and those without summaries as additional data for the camera trajectory generation. 

\paragraph{Contextual Analysis of the Dataset Descriptions.}
\begin{table*}[htbp]
    \centering
    \definecolor{darkgray176}{RGB}{176,176,176}
\definecolor{skyblue}{RGB}{135,206,235}
\definecolor{steelblue31119180}{RGB}{31,119,180}

\begin{tikzpicture}
\begin{groupplot}[
    group style={
        group size=2 by 2, 
        horizontal sep=2cm, 
        vertical sep=2cm    
    },
    height=5cm, 
    width=8cm,  
    xlabel={X-axis label}, 
    ylabel={Y-axis label}, 
]

\nextgroupplot[
    title={Number of Houses by Number of Floors},
    xlabel={Number of Floors},
    ylabel={Number of Houses},
    ymajorgrids,
    x grid style={darkgray176},
    y grid style={darkgray176},
    ymin=0,
    ymax=391.65,
    xmin=0.41,
    xmax=4.59
]
\definecolor{darkgray176}{RGB}{176,176,176}
\draw[draw=none,fill=steelblue31119180] (axis cs:1.6,0) rectangle (axis cs:2.4,373);
\draw[draw=none,fill=steelblue31119180] (axis cs:2.6,0) rectangle (axis cs:3.4,362);
\draw[draw=none,fill=steelblue31119180] (axis cs:0.6,0) rectangle (axis cs:1.4,63);
\draw[draw=none,fill=steelblue31119180] (axis cs:3.6,0) rectangle (axis cs:4.4,10);

\nextgroupplot[
    title={Number of Houses by Number of Bedrooms},
    xlabel={Number of Bedrooms},
    ylabel={Number of Houses},
    ymajorgrids,
    x grid style={darkgray176},
    y grid style={darkgray176},
    ymin=0,
    ymax=318.15,
    xmin=-0.89,
    xmax=9.89
]
\definecolor{darkgray176}{RGB}{176,176,176}
\draw[draw=none,fill=steelblue31119180] (axis cs:3.6,0) rectangle (axis cs:4.4,303);
\draw[draw=none,fill=steelblue31119180] (axis cs:2.6,0) rectangle (axis cs:3.4,258);
\draw[draw=none,fill=steelblue31119180] (axis cs:4.6,0) rectangle (axis cs:5.4,133);
\draw[draw=none,fill=steelblue31119180] (axis cs:1.6,0) rectangle (axis cs:2.4,47);
\draw[draw=none,fill=steelblue31119180] (axis cs:5.6,0) rectangle (axis cs:6.4,36);
\draw[draw=none,fill=steelblue31119180] (axis cs:0.6,0) rectangle (axis cs:1.4,17);
\draw[draw=none,fill=steelblue31119180] (axis cs:-0.4,0) rectangle (axis cs:0.4,7);
\draw[draw=none,fill=steelblue31119180] (axis cs:6.6,0) rectangle (axis cs:7.4,4);
\draw[draw=none,fill=steelblue31119180] (axis cs:7.6,0) rectangle (axis cs:8.4,2);
\draw[draw=none,fill=steelblue31119180] (axis cs:8.6,0) rectangle (axis cs:9.4,1);

\nextgroupplot[
    title={Number of Houses by Architecture},
    xlabel={Number of Bedrooms},
    ylabel={Number of Houses},
    ymajorgrids,
    xtick style={color=black},
    xtick={0,1,2,3,4,5,6,7,8,9,10,11,12},
    xticklabel style={rotate=45.0, anchor=east},
    xticklabels={
      modern,
      contemporary,
      colonial,
      victorian,
      traditional,
      cottage-like,
      rustic,
      cottage,
      arts and crafts,
      country,
      country-style,
      classic,
      century
    },
    x grid style={darkgray176},
    y grid style={darkgray176},
    ymin=0,
    ymax=550.2,
    xmin=-1.04,
    xmax=13.04
]
\definecolor{darkgray176}{RGB}{176,176,176}
\draw[draw=none,fill=steelblue31119180] (axis cs:-0.4,0) rectangle (axis cs:0.4,524);
\draw[draw=none,fill=steelblue31119180] (axis cs:0.6,0) rectangle (axis cs:1.4,261);
\draw[draw=none,fill=steelblue31119180] (axis cs:1.6,0) rectangle (axis cs:2.4,5);
\draw[draw=none,fill=steelblue31119180] (axis cs:2.6,0) rectangle (axis cs:3.4,5);
\draw[draw=none,fill=steelblue31119180] (axis cs:3.6,0) rectangle (axis cs:4.4,3);
\draw[draw=none,fill=steelblue31119180] (axis cs:4.6,0) rectangle (axis cs:5.4,2);
\draw[draw=none,fill=steelblue31119180] (axis cs:5.6,0) rectangle (axis cs:6.4,2);
\draw[draw=none,fill=steelblue31119180] (axis cs:6.6,0) rectangle (axis cs:7.4,1);
\draw[draw=none,fill=steelblue31119180] (axis cs:7.6,0) rectangle (axis cs:8.4,1);
\draw[draw=none,fill=steelblue31119180] (axis cs:8.6,0) rectangle (axis cs:9.4,1);
\draw[draw=none,fill=steelblue31119180] (axis cs:9.6,0) rectangle (axis cs:10.4,1);
\draw[draw=none,fill=steelblue31119180] (axis cs:10.6,0) rectangle (axis cs:11.4,1);
\draw[draw=none,fill=steelblue31119180] (axis cs:11.6,0) rectangle (axis cs:12.4,1);

\nextgroupplot[
    title={Number of Houses by Architecture},
    xlabel={Number of Bedrooms},
    ylabel={Number of Houses},
    ymajorgrids,
    xtick style={color=black},
    xtick={0,1,2,3,4,5,6,7,8,9},
    xticklabel style={rotate=45.0, anchor=east},
    xticklabels={
      detached,
      condo,
      row house,
      bungalow,
      semi-detached,
      cottage,
      commercial,
      office,
      loft,
      townhouse
    },
    x grid style={darkgray176},
    y grid style={darkgray176},
    ymin=0,
    ymax=717.15,
    xmin=-0.89,
    xmax=9.89
]
\definecolor{darkgray176}{RGB}{176,176,176}
\draw[draw=none,fill=steelblue31119180] (axis cs:-0.4,0) rectangle (axis cs:0.4,683);
\draw[draw=none,fill=steelblue31119180] (axis cs:0.6,0) rectangle (axis cs:1.4,45);
\draw[draw=none,fill=steelblue31119180] (axis cs:1.6,0) rectangle (axis cs:2.4,39);
\draw[draw=none,fill=steelblue31119180] (axis cs:2.6,0) rectangle (axis cs:3.4,14);
\draw[draw=none,fill=steelblue31119180] (axis cs:3.6,0) rectangle (axis cs:4.4,13);
\draw[draw=none,fill=steelblue31119180] (axis cs:4.6,0) rectangle (axis cs:5.4,8);
\draw[draw=none,fill=steelblue31119180] (axis cs:5.6,0) rectangle (axis cs:6.4,2);
\draw[draw=none,fill=steelblue31119180] (axis cs:6.6,0) rectangle (axis cs:7.4,2);
\draw[draw=none,fill=steelblue31119180] (axis cs:7.6,0) rectangle (axis cs:8.4,1);
\draw[draw=none,fill=steelblue31119180] (axis cs:8.6,0) rectangle (axis cs:9.4,1);

\end{groupplot}
\end{tikzpicture}
    \caption{\textbf{Contextual dataset statistics.} As shown in the data, most of the properties are modern, detached, with 2-3 floors, and 3-4 bedrooms.}
    \label{fig:context_analysis}
\end{table*}
The videos primarily feature tours of detached houses rather than flats or apartments, with most properties located in a city within a developed, financially advanced country and its surrounding suburbs (we refrain from disclosing the location for privacy reasons). The showcased real estate is predominantly high-end, often including luxurious features. As shown in Table \ref{fig:context_analysis}, these homes typically have three to five, with four being the most common. A substantial portion are multi-storey, usually with two or three floors, and many include outdoor spaces such as front or backyards. The interior design is largely modern or contemporary. It is important to note that our dataset is biased toward upscale properties in a specific region and does not capture the full diversity of global architectural styles.

\paragraph{Linguistic Analysis of the Dataset Descriptions.}

\begin{table*}[htbp]
\centering
\includegraphics[width=\textwidth]{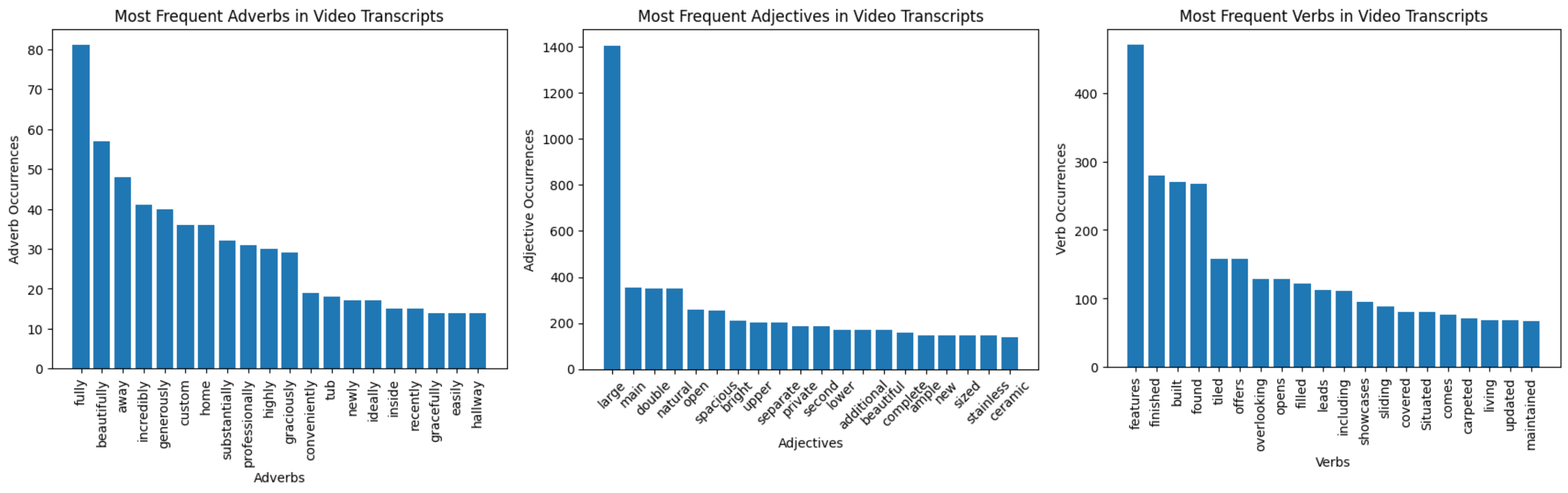}
\caption{\textbf{Constituency-based word frequency statistics.}}
\label{fig:constituency_stats}
\end{table*}

Table \ref{fig:constituency_stats} presents an analysis of phrases based on constituency within the extracted descriptions. The constituency-based analysis reveals that the most common adjectives in the descriptions fall into categories such as scalar (e.g., ``large", ``double", and ``ample"), directional (e.g., ``main", ``upper", and ``lower"), and conceptual (e.g., ``natural", ``open", and ``spacious"), all of which describe aspects of interior spaces. Additionally, there are adjectives related to material information, like ``stainless" and ``ceramic".

When analyzing the most frequent adverbs in the video descriptions, the word ``fully" stands out as the most frequent adverb, occurring approximately 80 times, which may suggest these videos often showcase homes that are fully equipped or fully furnished. Following ``fully," we see a significant mention of ``beautifully," which indicates a focus on aesthetic appeal, highlighting beautifully designed spaces. Other adverbs like ``away", ``incredibly", and ``graciously", suggest descriptions of location, extraordinary features, or hospitality aspects. Adverbs such as ``professionally", ``highly", and ``conveniently" point towards emphasizing quality and ease of living. The occurrence of these specific adverbs suggests that house tour videos prioritize aspects such as completeness, beauty, functionality, and unique features to attract potential buyers or viewers.

Lastly, the verbs within the descriptions give further linguistic cues on  descriptions as a whole. The verb ``features" appears most frequently, over 400 times, indicating a focus on highlighting key aspects or amenities of properties. Verbs like ``finished", ``built", and ``found" suggest emphasis on quality, construction, and location. Words such as ``offered", ``opens", and ``overlooking" reflect the dynamic aspects and views these properties provide. Other verbs like ``leads", ``includes", and ``showcases" emphasize navigation, inclusivity, and presentation within the space. The use of ``situated", ``covered", and ``updated" implies a focus on positioning, protection, and modern enhancements. Overall, these verbs underline the importance of showcasing distinctive features and conveying a sense of completeness and modernization in house tours.

\paragraph{Named-Entity Based Analysis of the Dataset Descriptions.}
\begin{table*}[htbp]
\centering
\includegraphics[width=\textwidth]{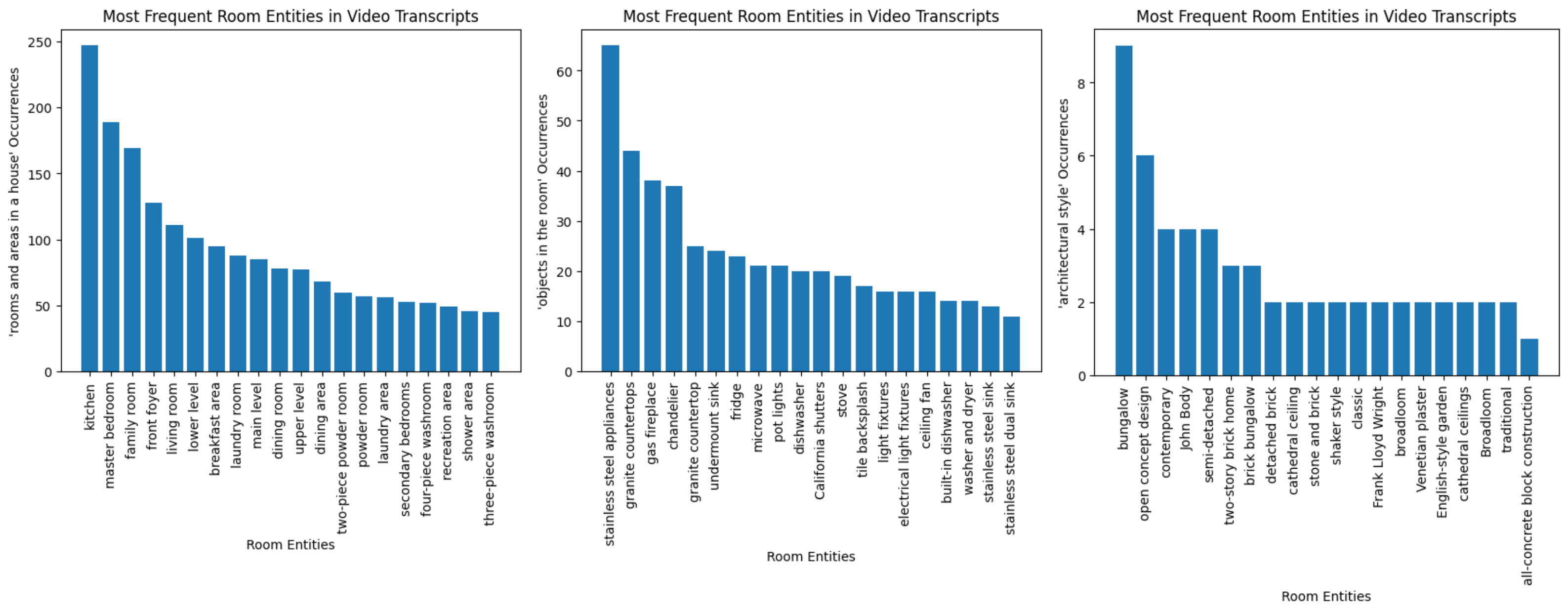}
\caption{Named Entity Frequency Statistics}
\label{fig:ner_stats}
\end{table*}
When analyzing the first bar plot in Table \ref{fig:ner_stats} the \textit{rooms and areas in a house} entities in video descriptions, the ``kitchen" emerges as the most highlighted space, with nearly 250 mentions, underscoring its role as a central and significant feature in homes. The ``master bedroom" follows closely, reflecting its importance in personal comfort and privacy. Other commonly referenced areas include the ``family room," ``front foyer," and ``living room," which are key spaces for gathering and welcoming guests. The plot also shows notable mentions of functional areas like the ``laundry room" and ``dining room," indicating their relevance in daily living. The presence of terms like ``lower level" and ``breakfast area" suggests an emphasis on specific sections or niches that add value to the property's layout. Overall, the focus on both communal and private spaces highlights a balanced presentation of essential living areas and unique features in house tours.\newline

The second plot shows the frequency of \textit{objects in a room}. ``Stainless steel appliances" are the most prominent, appearing over 60 times, emphasizing the modern and desirable features of kitchens. Following closely are ``granite countertops" and ``gas fireplaces," indicating a focus on quality materials and cozy elements. Decorative items like ``chandeliers" highlight style and luxury in living spaces. Functional features such as ``undermount sinks," ``fridges," and ``microwaves" underscore practical aspects of home living. The presence of ``pot lights" and ``California shutters" suggests attention to lighting and window treatments. Additional mentions of ``stoves," ``dishwashers," and ``tile backsplashes" reflect both essential and aesthetic components of kitchens.\newline

The third and the last plot visualizes the \textit{architectural style} found within the real estates. ``Bungalow" emerges as the most frequently cited style, followed by ``open concept design," indicating a preference for single-storey (usually plus the lower floor) living and spacious, flowing interiors. Styles like ``contemporary" and ``semi-detached" also receive notable mentions, highlighting a mix of modern and practical designs. The inclusion of ``two-storey brick home" and ``brick bungalow" suggests an appreciation for classic, durable architecture. Terms like ``cathedral ceiling" and ``stone and brick" emphasize distinct design features and materials. Styles attributed to iconic architecural figures like ``Frank Lloyd Wright" reflect a nod to renowned architectural influences. The mentions of ``English-style garden" and ``Venetian plaster" suggest an interest in incorporating thematic and textural elements. Overall, this plot underscores a variety of styles, balancing traditional and modern influences in architectural preferences.

\begin{figure*}[h!]
    \includegraphics[width=\linewidth]{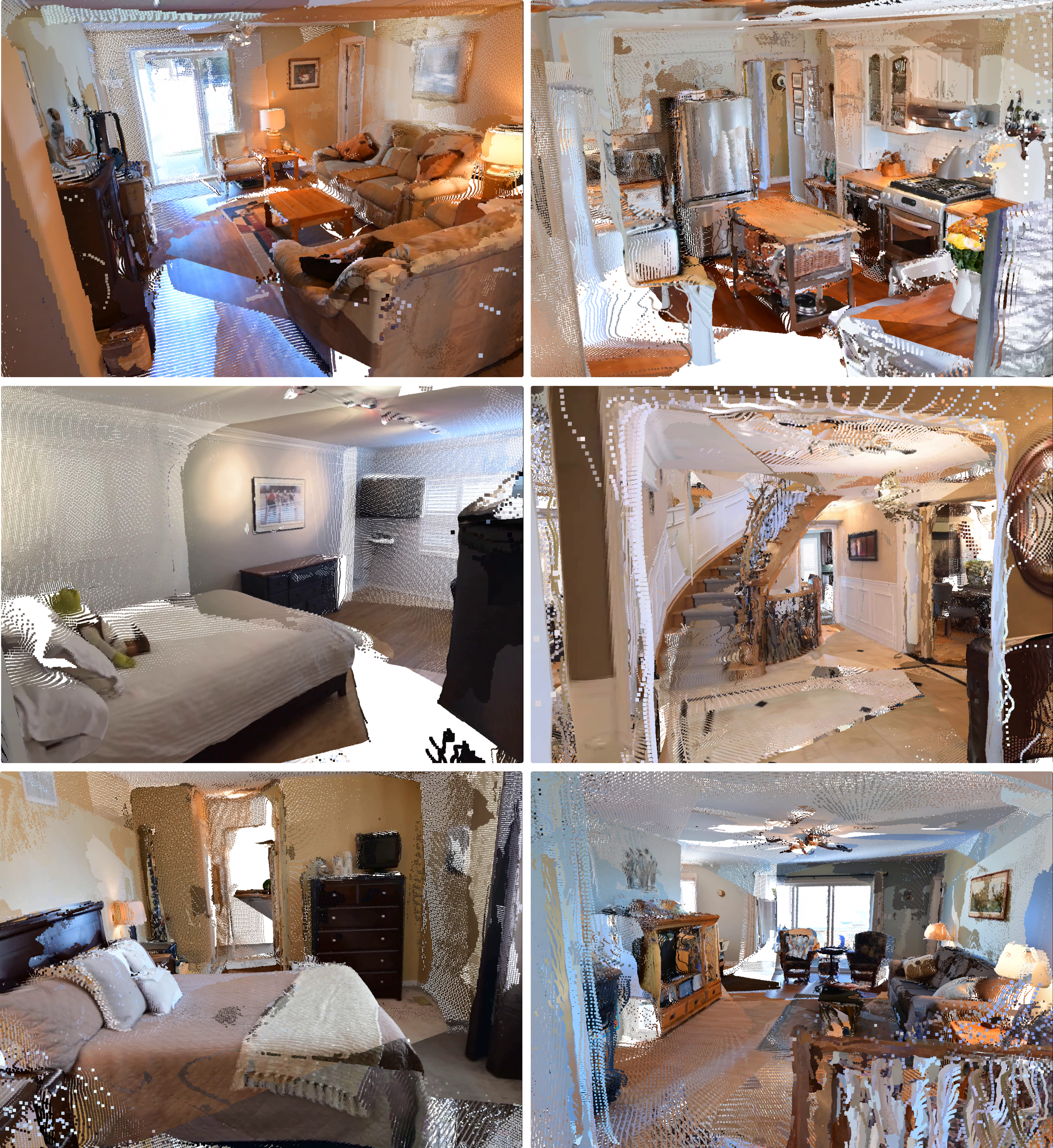}
    \captionof{figure}{\textbf{Gallery of sample captures from the \project dataset}. The showcased regions originate from the 3D reconstructions.}\label{fig:dataset_gallery}
\end{figure*}

\paragraph{Captures from the \project{} Dataset.}
In Figure \ref{fig:dataset_gallery}, we provide some captures from the dense point clouds of our \project{} dataset.

\section{Additional Qualitative Results}
\label{sec:supp_quals}
\begin{figure*}[htp]
        \includegraphics[width=\textwidth]{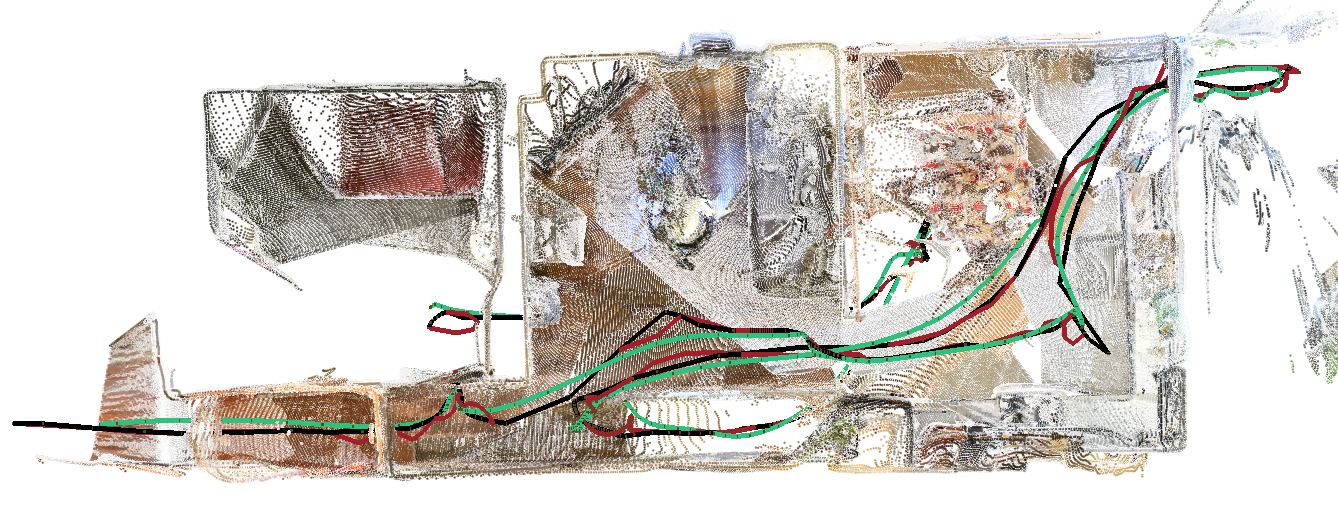}%
    \caption{\textbf{Trajectory Visualization Within the 3D Reconstructions (top view).} Our method, Residual Diffuser, achieves a more human-like and smooth trajectory than the baseline Catmull-Rom Spline. \textbf{Black:} Ground-Truth, \textbf{Green:} \textit{Residual Diffuser} and \textbf{Red:} Catmull-Rom Spline.}
    \label{fig:supp_trajquals}
\end{figure*}

\begin{figure*}[h!]
    \includegraphics[width=\linewidth]{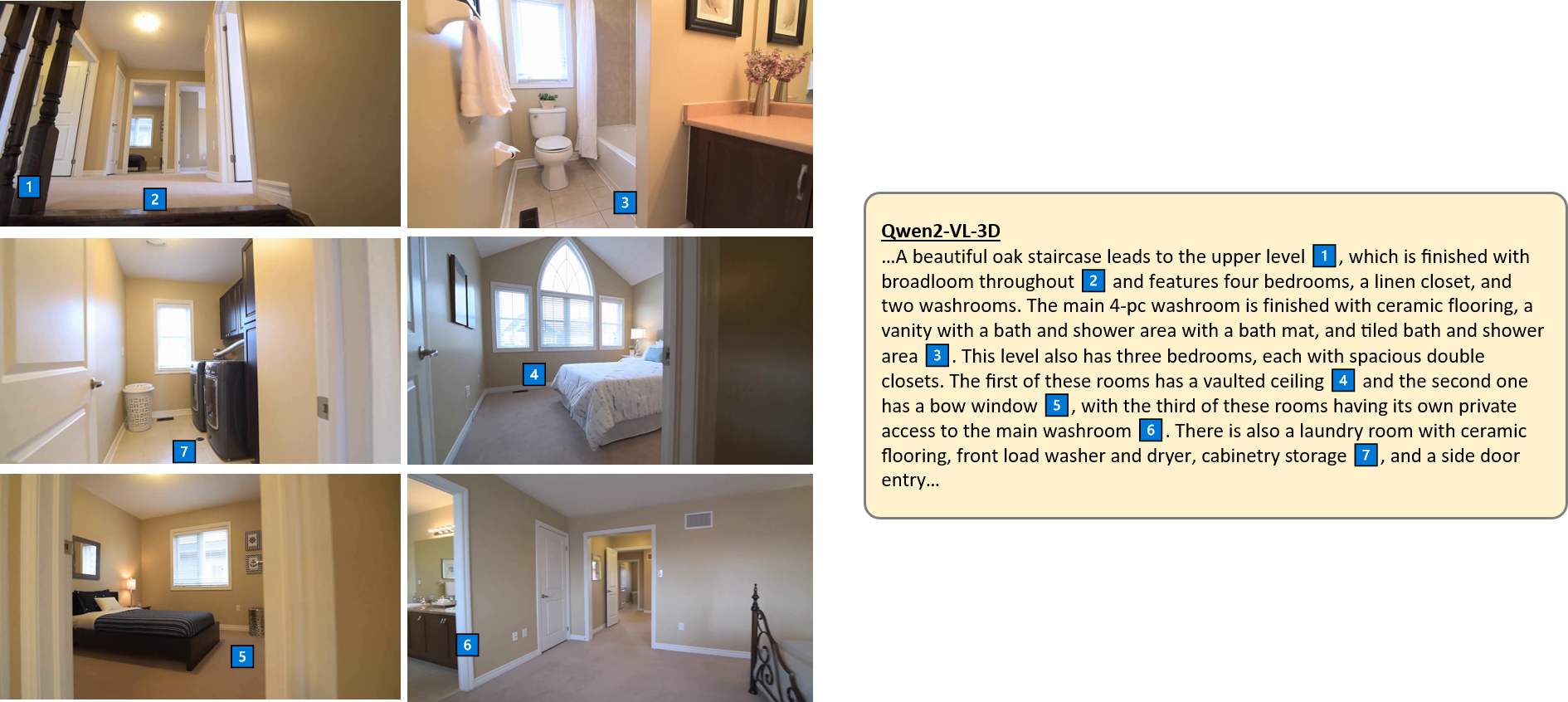}
    \captionof{figure}{Further qualitative results for scene-level summary generation}\label{fig:summary_quals_2}
\end{figure*}

Figure \ref{fig:supp_trajquals} provides additional qualitative results on trajectory generation from \textit{Residual Diffuser}, and Figure \ref{fig:summary_quals_2} on scene-level summary generation from \textit{Qwen2-VL-3D}.   

\section{User Evaluation of Text Generation}
\label{sec:user_eval}

\begin{table}[htbp!]
\centering
\resizebox{\columnwidth}{!}{
\begin{tabular}{cccccc} 
\toprule
\textbf{Methods} & \textbf{Lay.} & \textbf{Mat.} & \textbf{Fix.} & \textbf{Amb.} & \textbf{Ove.} \\
\midrule
Qwen2-VL-7B (SFT) & 6.1 & 6.0 & 5.9 & 5.9 & 6.0 \\
Qwen2-VL-3D (Ours) & \cellcolor{green}\textbf{7.0} & \cellcolor{green}\textbf{7.3} & \cellcolor{green}\textbf{7.2} & \cellcolor{green}\textbf{6.9} & \cellcolor{green}\textbf{7.3} \\
\bottomrule
\end{tabular}
}

\caption{\textbf{User Study} evaluating text generation quality across five categories: \textbf{Layout} (Lay.), \textbf{Material} (Mat.), \textbf{Fixture} (Fix.), \textbf{Ambience} (Amb.), and \textbf{Overall} (Ove.). Score range: [0,10].}

\label{tab:user_study}
\end{table}

We conducted a single-blind user study in which three different participants evaluated generated descriptions for 20 different scenes. Across all assessed categories (Tab. \ref{tab:user_study}), users consistently preferred our method. The results suggest that incorporating 3D positional information significantly enhances the perceived quality of the generated text.
In our experimental setup, each participant watched a house tour video accompanied by two textual descriptions: one generated by a standard fine-tuned (SFT) baseline and the other by our Qwen2-VL-3D model. For every video, the order of the two descriptions was randomized to eliminate order bias. Six participants rated each description on a 0–10 scale, where 0 indicates no correspondence with the video and 10 indicates perfect alignment. The two descriptions were shown simultaneously, allowing for both absolute and comparative assessments. The definitions for each grading category is as follows:\\
\noindent\textbf{Layout.} Does the description demonstrate a good understanding of the spatial organization of the room? Consider references to walls, doors, windows, room shapes, and how space is structured.\\
\noindent\textbf{Material.} Does the description accurately capture the materials and finishes in the scene? Look for details about surface textures, colors, or types of materials (e.g., wood, glass, concrete).\\
\noindent\textbf{Fixture.} How well does the description mention relevant built-in or fixed elements? Examples include lighting fixtures, sinks, cabinetry, or any permanent installations.\\
\noindent\textbf{Ambiance.} Does the description convey the overall mood or atmosphere of the space? Consider lighting, color tone, and emotional or sensory impressions.\\
\noindent\textbf{Overall.} An overall grading on the quality of the description.

\section{Implementation Details}
\label{sec:supp_implement}

\textbf{Residual Diffuser.} We employ a lightweight U-Net architecture with two downsampling and two upsampling layers, changing the trajectory length by a factor of two at each step. The model is trained on a single \textit{NVIDIA GeForce RTX 2080 (8GB)} GPU for 30K iterations with a batch size of 1 and gradient accumulation every 8th iteration, using a learning rate of $5 \times 10^{-6}$. To improve generalization, we randomly vary the number of sparse observations per training step, ensuring at least one observation every 20 frames.

\noindent\textbf{Qwen2-VL-3D.} We adopt a two‐step training strategy for Qwen2‐VL‐3D. In the first phase, we LoRA‐finetune Qwen2‐VL on a single \textit{NVIDIA A100 (80GB)} GPU for 20 epochs, with early stopping after the 8th epoch. We use the AdamW optimizer with a learning rate of \(2 \times 10^{-4}\), cosine scheduling, and a warm-up ratio of 0.03. The effective batch size is 1, with gradient accumulation every 8th iteration. We train the VLM in \textit{bfloat16} precision and apply gradient clipping at a maximum norm of 0.3 to prevent overflow. For the LoRA adapter, we specify both rank and alpha as 64, along with a 0.05 dropout rate. Adapters are added only to the Q and V weights of the attention layers.
\newline
\noindent\textbf{Inference Times.} All benchmarks were run on a 40 GB NVIDIA A100 GPU. The lightweight Residual Diffuser generates one trajectory in 0.23±0.04 s on average, whereas Qwen2-VL-3D requires 14.9±5.6 s to produce a single textual response.

\section{Out-Of-Distribution Scenes}
\label{sec:ood_scenes}

\begin{figure*}[htbp]
    \centering
    \includegraphics[width=1 \textwidth]{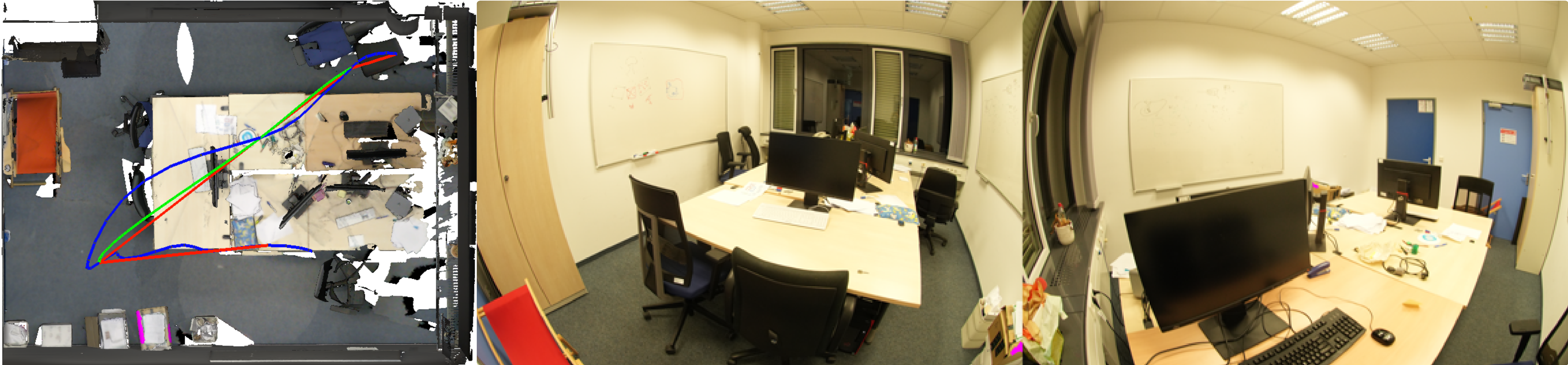}
    \begin{tcolorbox}[
      before skip=0pt,
      after skip=0pt,
      colback=yellow!40,   
      colframe=black!50,   
      boxrule=0.5pt,       
      arc=1pt,             
      left=1pt, right=1pt, top=1pt, bottom=1pt,
      width=\textwidth,
      fontupper=\small
    ]
    ``The spacious office features a large window with blinds for privacy and light control, as well as a whiteboard on the opposite wall for brainstorming sessions or team meetings. The open layout includes multiple workstations equipped with ergonomic chairs, computer monitors, and desktops. There is also a large desk area with additional chairs for collaborative workspaces. The office has ample storage options, including a wooden file cabinet, a credenza, and several deskside drawers. A large closet provides additional storage space, while a small closet near the door offers personal belongings storage. This room would be suitable for a variety of professional settings, such as law firms, accounting offices, architectural firms, engineering firms, and more."
    \end{tcolorbox}
    \caption{\textbf{Inference on ScanNet++.} Generated text and trajectory based on 5 frames from scene: \textit{56a0ec536c}.}
    \label{fig:scannet_capt}
\end{figure*}

\begin{figure}[htbp]
    \begin{tcolorbox}[
      colback=yellow!40,   
      colframe=black!50,   
      boxrule=0.5pt,       
      arc=1pt,             
      left=1pt, right=1pt, top=1pt, bottom=1pt,
      width=\columnwidth,
      fontupper=\small
    ]
    ``The home features a large front porch with a decorative ceiling and columns, as well as a front yard with a landscaped garden bed. The home has been freshly painted and features a new roof, new windows, and new flooring throughout. The home also includes a detached two-car garage with an attached carport, a large backyard with a garden bed, and a covered patio area."
    \end{tcolorbox}
    \vspace{-16pt}
    \caption{\textbf{Inference on online video.} Generated text based on exterior drone shot, \url{https://www.youtube.com/watch?v=QZrjZbI-H00}.}
    \label{fig:drone_capt}
\end{figure}

To evaluate the generalization capabilities of our method, we test it on two out-of-distribution (OOD) scenarios: (1) a single office room from the ScanNet++ dataset and (2) an online drone-view exterior shot of a property (Fig. \ref{fig:drone_capt}.
For these examples, we adjust the decoding temperature to $T=0.3$ (compared to $T=1.0$ used on the in-distribution HouseTour dataset). The temperature scaling sharpens the softmax distribution and, in practice, lets the weak image evidence outweigh generic language priors, dampening the language priors learned during training (e.g., persistent mention of bedrooms or kitchens in training samples); hence curtailing hallucinations that do not align with the visual evidence. While the tweak is effective, it also showcases the narrower visual-text diversity of the training data; expanding the dataset should further improve OOD robustness and lessen the reliance on temperature scaling. Note that, in both cases, the generated text is featuring the learned language style.
As shown in Figure \ref{fig:scannet_capt}, the trajectory we generate (shown in blue) exhibits smoother and more natural motion compared to linear interpolation (red) and Catmull-Rom splines (green).

\section{Evaluation In Metric Scale}
\label{sec:metric_eval}

\begin{figure}[htbp]
    \centering
    \begin{minipage}{0.48\linewidth}
        \subfloat[]{%
            \includegraphics[width=\linewidth]{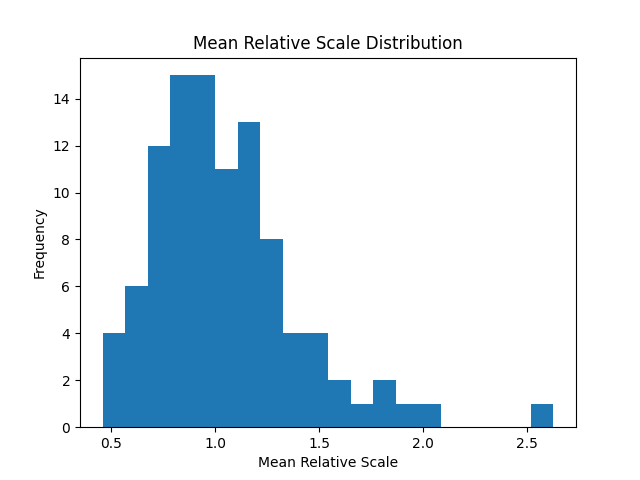}%
        }
    \end{minipage}
    \hfill
    \begin{minipage}{0.48\linewidth}
        \subfloat[]{%
            \includegraphics[width=\linewidth]{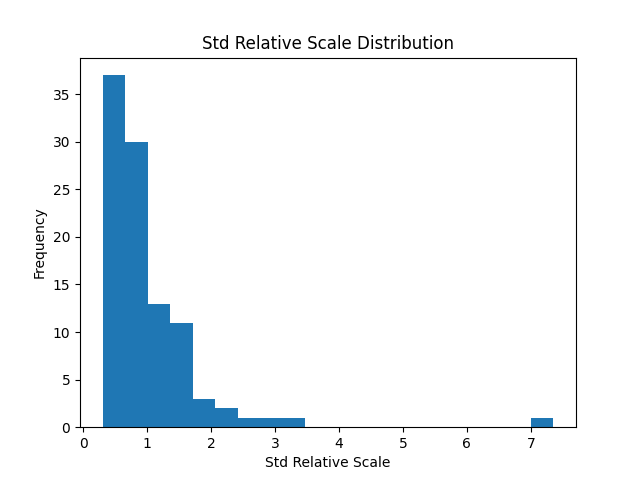}%
        }
    \end{minipage}
    \caption{The \textbf{Mean} and \textbf{Standard Deviation} of the scale multipliers required to achieve metric scale.}
    \label{fig:mean_std_metric}
\end{figure}

In our experiments, we report all evaluation results in metric scale, even though our 3D reconstructions are not inherently metric. We adjust the scale of the reconstructions for evaluation so that errors in both large and small scenes are handled consistently, avoiding the distortion that arises from varying relative scales.

To achieve this, we use the same metric depth model employed by Mast3r \cite{Leroy2024MASt3R} for metric alignment of camera pose pairs. Specifically, we uniformly sample 20 pairs of sequential keyframes from each reconstruction and measure the Euclidean distances between their poses as estimated by the Mast3r model. We then compare these distances with the corresponding Euclidean distances in our reconstructions. The ratio of the Mast3r-based distance to the reconstructed distance gives a scale multiplier for each pair, and we average these values at the scene level. The resulting mean scale multiplier for a scene is then applied to align that scene’s trajectory to metric scale.

Figure \ref{fig:mean_std_metric} shows the mean and standard deviation of the scene-wise metric scale multipliers. As indicated by the figure, most of the scale multipliers fall within a reasonable range, with low variance.

\section{Bradley-Terry Evaluation}
\label{sec:supp_BT}
In this section, we explain our evaluation process to generate Bradley-Terry (\textbf{BT}) normalized scores (between 0 and 1) for each of the Multi-Image-to-Text methods.

\noindent \paragraph{Algorithm.}
We begin by creating pairs of generated summaries, comparing outputs from each method to the ground truth for all 130 scenes in our test set. To mitigate positional bias, we shuffle these pairs to randomize the order in which they are presented to the LLM. For evaluation, we use the GPT-4o model \cite{hurst2024gpt} with a temperature of $0.5$ for text generation. Although we also tested open-source models like Llama-3.1-7b, we found they exhibit strong ordering bias, consistently favoring the first summary presented.

After gathering the binary preferences from the GPT-4o model, we construct a preference matrix $\mathcal{M}(i, j)$, where the entry in the $i^\text{th}$ row and $j^\text{th}$ column represents how many times method $i$ is preferred over method $j$. The score calculation algorithm starts with an initial guess for the parameters: each method is assigned a value of 1 (stored in the array $\pi_i$). These parameters represent the latent “strength” or ability of each method. We iteratively update the values of $\pi_i$ as follows:
\[
\pi_i = \frac{wins_i}{\sum_{j \neq i} \frac{N[i, j]}{\pi_i + \pi_j}},
\]
where $wins_i$ is the total number of wins for method $i$, and $N[i, j]$ is the total number of comparisons between methods $i$ and $j$. The denominator arises from setting the derivative of the likelihood function to zero (maximizing the likelihood), and it reflects how often method $i$ has been compared with method $j$ relative to their current estimated abilities. After updating all the parameters, the algorithm checks whether the maximum change in any parameter is below a specified tolerance threshold. If so, the iteration stops, indicating convergence.

\noindent \paragraph{Transforming Parameters to Scores.}
Once the parameters $\pi$ have been estimated, they are converted into scores $s_i$ in the interval $(0, 1)$ using the logistic transformation:
\[
s_i = \frac{\pi_i}{\pi_i + 1}.
\]
This transformation is useful because it maps the potentially unbounded $\pi_i$ values to probability-like scores, making them easier to interpret.

\noindent \paragraph{Prompt to the Judge.}
We use the following prompt as input to the GPT-4o judge:
\\\\
\texttt{"Given Prediction 0, Prediction 1 and the Ground-Truth texts, select which text is closer to the ground-truth. Evaluate the texts only based on the information available in ground-truth."}

\section{Recall Curves}
\label{sec:supp_recall}
\begin{figure}[htbp]
    \includegraphics[width=\columnwidth]{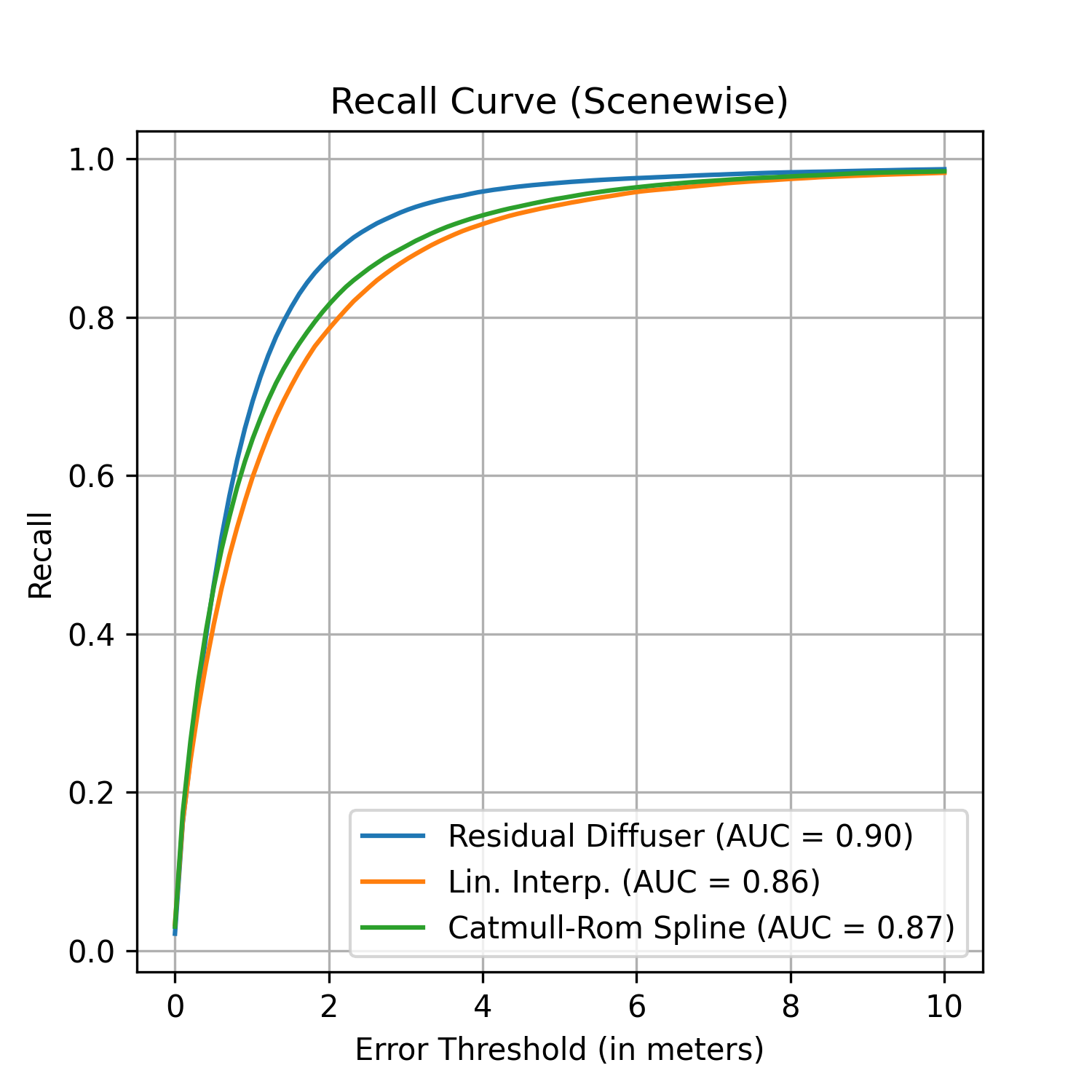}
    \captionof{figure}{\textbf{Recall curve for 3D camera trajectory generation.} The x-axis shows the error threshold (in meters), and the y-axis indicates the ratio of predictions with errors below this threshold. }\label{fig:recall_curve}
\end{figure}
In our trajectory generation evaluation, we use recall as a way of quantifying the magnitude of errors that the generation methods achieve. In the main paper tables (Tables 1 and 2) we report the results for \textit{R@50cm}, \textit{R@75cm} and \textit{R@1m}. We provide the complete curves in Figure \ref{fig:recall_curve}. Our method is shown to have the highest Area Under Curve (\textbf{AUC}) score and consistently shows better performance against the baselines with varying error thresholds.

\section{Preliminaries}
\label{sec:supp_prelim}
 Generative modeling using denoising diffusion probabilistic models (DDPMs) aims to learn a probability distribution $p_{\theta}(\mathbf{x})$ that approximates the true data distribution of observed data $\mathbf{x}$. Unlike other generative methods -- such as variational autoencoders or generative adversarial networks -- which generate data in a single step, DDPMs gradually transform pure noise into structured data through an iterative denoising process. The discrete stochastic denoising (reverse) process is modeled as a Markov chain, beginning at a predefined time step $T$ where the signal is considered to be pure noise $p(x_T) = \mathcal{N}(x_T; 0, I)$. A neural network $\epsilon_\theta$ is trained to predict the noise added at each timestep by minimizing the variational bound on the negative log likelihood, $\mathbb{E}[-log(p_\theta(x_0))]$. In practice, the reverse process is typically parametrized using Gaussian distribution as follows: 
\begin{equation}
\label{eq:diff_reverse}
    p_\theta(x_{t-1}|x_t) = \mathcal{N}(x_{t-1}; \mu_\theta(x_t,t), \Sigma_\theta(x_t, t)) 
\end{equation}
Forward diffusion is a process that gradually adds noise to the data via a variance schedule $\beta_t \in (0, 1)$, determining the amount of noise introduced at each timestep $t$. This formulation enables a closed-form expression for sampling an arbitrary $x_t$, where $\alpha_t := 1 - \beta_t$ and $\bar{\alpha}_t := \prod_{0 \leq i \leq t} \alpha_i$. The conditional probability distribution $q(x_t|x_0)$ describes how likely $x_t$ is, given the clean signal $x_0$:
\begin{equation}
\label{eq:diff_forward}
    q(x_t|x_0) = \mathcal{N}(x_t; \sqrt{\bar{\alpha_t}}x_0, (1 - \bar{\alpha_t}) I)
\end{equation}

\end{document}